\documentclass{article}
\PassOptionsToPackage{numbers,sort&compress}{natbib}
\usepackage[eandd,preprint]{neurips_2026}

\usepackage[utf8]{inputenc}
\DeclareUnicodeCharacter{2212}{\ensuremath{-}}
\DeclareUnicodeCharacter{03C1}{\ensuremath{\rho}}
\DeclareUnicodeCharacter{03C4}{\ensuremath{\tau}}
\DeclareUnicodeCharacter{2032}{\ensuremath{{}^{\prime}}}
\usepackage[T1]{fontenc}
\usepackage{hyperref}
\usepackage{url}
\usepackage{booktabs}
\usepackage{amsfonts}
\usepackage{amsmath}
\usepackage{nicefrac}
\usepackage{microtype}
\usepackage{xcolor}
\usepackage{graphicx}
\usepackage{multirow}
\usepackage{array}
\usepackage{longtable}
\usepackage{calc}
\usepackage{placeins}
\usepackage{float}
\usepackage{pifont}
\usepackage[table]{xcolor}

\newcolumntype{L}[1]{>{\raggedright\arraybackslash}p{#1}}
\newcolumntype{C}[1]{>{\centering\arraybackslash}p{#1}}

\graphicspath{{figures/}}
\hypersetup{hidelinks,hypertexnames=false}

\title{TrialAgentBench: Evaluating AI Agents for Clinical-Trial Analysis and Long-Horizon Drug-Development Decisions}

\author{Bradley Segal\\University of Oxford\\\texttt{bradley.segal@wolfson.ox.ac.uk} \And William Bolton\\University of Oxford\\\texttt{will.bolton@eng.ox.ac.uk} \And Philip Torr\\University of Oxford\\\texttt{philip.torr@eng.ox.ac.uk} \AND David Clifton\\University of Oxford\\\texttt{david.clifton@eng.ox.ac.uk} \And Lei Clifton\\University of Oxford\\\texttt{lei.clifton@phc.ox.ac.uk}}
\hypersetup{pdftitle={TrialAgentBench: Evaluating AI Agents for Clinical-Trial Analysis and Long-Horizon Drug-Development Decisions},pdfauthor={Bradley Segal, William Bolton, Philip Torr, David Clifton, Lei Clifton}}

\begin{document}
\raggedbottom

\maketitle

\begin{abstract}
Clinical drug development is a long-horizon, high-stakes process requiring sequential decisions based on information that costs money and time to acquire. AI agents have demonstrated impressive capabilities for scientific analysis, but there is no established way to evaluate them on these decisions, as real trials are expensive, involve people, and never reveal the counterfactual. We introduce TrialAgentBench, a benchmark for evaluating AI agents on clinical trial analysis and drug development decision-making using simulated trials where the truth is known, embedded within the data records of a fictional drug sponsor. We publicly release nearly 3,500 tasks, varying from analysing a single trial to managing multi-phase programmes with budget and time constraints. To separate how well agents read evidence from how well they act on it, we score each analysis by rerunning its code on new trials, and each programme by the value its decisions achieve. Across 24 closed- and open-weight models, the strongest agents can now analyse raw, messy trial data records to within 6\% of a process-aware analysis, even when an assumption is violated, with 95\% confidence intervals containing the true effect 94--95\% of the time. Over the prior year, the best model improved from 19\% above this analysis, with serious misses on 15\% of tasks, to 3.5\% and 1\%. What limits them is decision-making from evidence under uncertainty: when to correct, what evidence to buy, when to stop. The strongest agents are overly careful, correcting on 83\% of tasks with equal recording errors, where it is unnecessary. On drug programmes, the best agent has substantial headroom, reaching 48\% of the value an oracle achieves against 58\% for a rule-based developer, with others falling 24 to 66 points short of that developer. Model generations show ongoing improvements in handling data and evidence, with substantial variation. The strongest models can act as competent analysts, but do not yet make reliable long-horizon decisions.
\end{abstract}

\section{Introduction}

New medicines are one of the most direct ways that science can improve people's lives. Every medicine that makes it to patients has the potential to cure a disease, slow its progression, or spare a patient from a worse treatment. The hope is that clinical trials can be made faster, cheaper, and better, so that more medicines can reach patients sooner and with stronger evidence. AI agents could help us get there. The path to developing a new drug is long and costly: including failures, the median cost to develop a drug is US\$1.1 billion \cite{wouters2020estimated}, and most drug candidates that enter clinical trials are never approved \cite{wong2019estimation}. Every decision along the way has to balance time and money against information. A pilot study may lead to a sponsor abandoning a promising drug or designing an underpowered follow-up study \cite{kraemer2006caution}, and a small, borderline result can just as easily commit a sponsor to years of work. In amyotrophic lateral sclerosis, a phase 2 study of dexpramipexole that was not powered for efficacy compared two doses in 92 participants in its second part and reached P = 0.046, which its authors said strongly supported further testing. The phase 3 trial that followed enrolled 943 participants and showed no benefit over placebo \cite{cudkowicz2011dexpramipexole, cudkowicz2013empower}.

Drug development is a multi-phase process, where each step is designed based on the results of the previous one. The results of a safety study determine the regimens tested next, the results of an exploration study determine the most promising candidate to advance, and its estimated effect guides the size of the subsequent confirmation trial, which may take years. Obtaining accurate results requires aligning trial data with the study protocol, handling participant dropout and treatment switching, considering the trial design, and verifying the quality of site records. An analysis that gets these wrong can still produce a precise estimate with a narrow interval. If the effect is overestimated by one standard error, the follow-up trial will be about half the size it needs to be, reducing the chance of detecting a true benefit from 80\% to about 54\%, so a drug that works may fail after years of development. If the effect is underestimated by one standard error, the follow-up trial will be over twice as large as needed, and the cost may lead a sponsor to abandon a drug that works.

AI agents have demonstrated strong performance in writing code and analysing data. But can we trust them to perform these analyses and make the decisions that depend on them? Real-world data cannot answer this question, because the true effect is never observed. Reanalysing published clinical trials can lead to different conclusions \cite{ebrahim2014reanalyses}, and different analysts can produce different estimates from the same dataset \cite{silberzahn2018many}. An evaluation on real data can test whether agents can replicate the original analysis, flaws included, but not whether they can recover the truth. Statisticians have long faced this challenge when developing new methods, and address it by evaluating their methods on simulated data where the true effect is known \cite{morris2019simulation, dorie2019automated}. Regulators such as the FDA similarly use simulation to evaluate complex clinical trial designs \cite{fda2020cid}.

TrialAgentBench brings this approach to AI agents, for both the analyses and the decisions that depend on them. In TrialEval, an agent is given a completed trial's documents and raw data records. It must align these records with the documents, analyse them, and save its analysis as a script, which we rerun on new trials. The 420 public tasks cover 17 clinical questions, from childbirth safety to metastatic cancer, across trial designs from individual randomisation to stepped-wedge rollouts and sizes from 52 to 114,000 participants. In TrialDev, an agent manages a drug development programme. Given 3 or 5 candidates, a budget of up to US\$314 million, and six years, it must decide which studies to purchase across three phases, from at least \(10^{21}\) combinations of study designs, read the results as they arrive, and decide when to stop and which drug to advance. Each of the 384 public programmes (3,072 tasks) is built on a simulated population of 64,000 participants. As for a real sponsor, a programme is worth the profit an approved drug generates over its market lifetime, minus the cost of harm, all studies, and each day it runs. We evaluate 24 models on TrialEval and 10 on TrialDev, and release the tasks, harness, replication trials, and an independent verifier.

\begin{figure}[t]\centering
\includegraphics[width=0.8\linewidth,height=0.9\textheight,keepaspectratio]{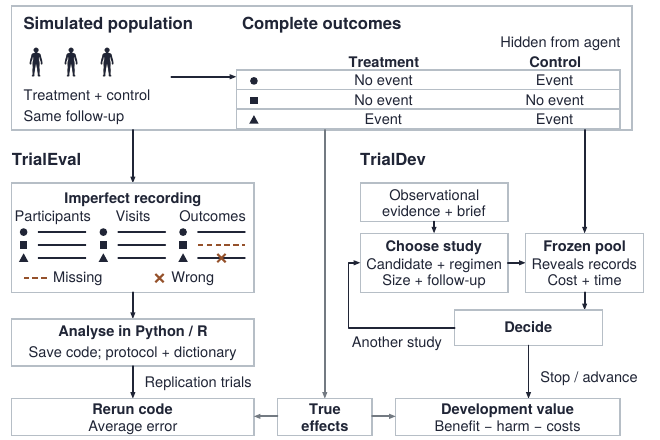}
\caption{TrialAgentBench scores trial analysis and development decisions against a simulated true effect hidden from the agent. Every participant has an outcome under each treatment. TrialEval supplies trial documents and records; the agent saves code, which is re-run on replication trials and scored by average error. TrialDev supplies a brief defining costs, approval rules and the value of benefit and harm. The agent buys studies from a frozen participant pool and is scored on its stop-or-advance decision.}
\end{figure}

\section{Related Work}

AI benchmarks have evolved to keep pace with the frontier of what models and agents can do to advance science. Recent benchmarks evaluate whether agents can resolve real-world software issues, write scientific code, and perform data-driven scientific discovery \cite{jimenez2024swebench, chen2024scienceagentbench, gu2024blade, tian2024scicode}; answer expert-validated questions \cite{rein2023gpqa}; demonstrate general and medical knowledge and clinical reasoning \cite{hendrycks2020measuring, liang2023holistic, singhal2023large, kanithi2024medic, arora2025healthbench, tang2025medagentsbench}; operate in virtual clinical environments \cite{jiang2025medagentbench}; and complete general agent, machine learning engineering, and data analysis tasks \cite{liu2023agentbench, ma2024agentboard, mialon2023gaia, huang2024mlagentbench, chan2024mlebench, hu2024infiagent, jing2024dsbench}.

Clinical trial benchmarks typically score against observed answers. Outcome prediction benchmarks score predictions of a trial's outcome made from its drug, target disease and eligibility criteria \cite{fu2021hint, chen2024trialbench} or forecast before its results are reported \cite{wang2026ctopen}. Others score how well models match patients to trials \cite{jin2023trialgpt}, find, screen and extract evidence as published reviews did \cite{wang2025trialpanorama, wang2025trialmind}, propose baseline participant characteristics to collect \cite{neehal2024ctbench}, judge whether published trials adhere to reporting guidelines \cite{he2025rctbench}, and reach the conclusions of systematic reviews \cite{polzak2025medevidence}. Closer to analysis, StatQA scores selection of the statistical methods that apply to a task \cite{zhu2024statqa}, QRData scores answers to statistical and causal questions based on data sheets from textbooks and academic papers \cite{liu2024qrdata}, BioDSBench scores code against test cases based on published biomedical analyses \cite{wang2026biodsbench}, and RWE-bench has agents execute a study protocol on a real-world database and scores how well the results match those published \cite{li2026rwebench}. For drug development strategy, one approach predicts the next clinical trials for an oncology programme and scores these against what the sponsor actually launched \cite{bolton2026clinicaltrialstrategy}. In almost all of these benchmarks, the answer is derived from a record: an expert's label or a published result. If our goal is to know that agents can reliably improve the quality of analyses, this is insufficient. An agent that aligns with the record may simply be learning to replicate human patterns, including their imperfections. To assess whether agents are advancing the field, we need to benchmark them against a simulated reality, where the answer is known and independent of analysts' judgement.

Statistics has long used simulated reality in this way: simulation studies and causal inference competitions compare results to a known answer \cite{morris2019simulation, dorie2019automated, hahn2019acic2017}. Causal reasoning benchmarks have extended this idea to language models \cite{jin2023cladder, jin2023corr2cause}, and recent agent benchmarks have extended it to data analysis. CausalDS scores answers on observational data generated from sampled causal models \cite{leban2026causalds}, while CauSciBench includes synthetic scenarios with a specified effect and scores both the method selected and the numerical answer against a reference \cite{acharya2026causcibench}. BoxingGym has agents choose experiments in known generative models \cite{gandhi2025boxinggym}. For long-horizon decision-making, Vending-Bench tasks agents with managing a virtual business \cite{backlund2025vendingbench, andon2026vendingbench2}. To our knowledge, no benchmark uses a simulated reality to test whether agents can be relied on for both trial analysis and drug development: analysing individual patient data from clinical trials, then managing information across a multi-year programme to maximise value under competing, real-world pressures like time and cost. Testing the same agents on both lets us follow them from reading a single trial to the decisions that bring a medicine to patients, and see where along that path they hold up and where they need to improve.

\section{TrialAgentBench}

TrialAgentBench runs on a clinical trial simulation compiler that translates trial specifications into programmable clinical trials. Given a specification, a copula-based model generates correlated participant features, a treatment assignment and disease progression model produces the outcome of each participant under each treatment, and an observation model produces the data that an analyst would receive. The true effect is therefore set before any data are collected. For example, if in a simulated population the 2-year risk of kidney failure is 18\% if everyone is on the drug and 23\% if everyone is on control, then the true effect is −5 percentage points, a quantity no real clinical trial can observe, as each participant is only seen under one condition. The true answer to a task also depends on the exact question. For example, two analysts can both be correct but disagree, if one estimates the effect of being assigned to the drug and the other estimates the effect of being on the drug. For a task to have a unique true answer, it needs to be clearly defined, including the population of interest, the treatment, the outcome, how to handle events such as treatment discontinuation, and the population-level summary metric. This is known as the estimand \cite{ich2019e9r1, cro2022estimands, kahan2024estimands}.

Each TrialEval task fixes the estimand in its analysis request, so every task has a single true effect against which any analysis can be scored. TrialEval captures the spectrum of clinical trials that a statistician would encounter in practice. The trials include individual and cluster randomisation, stepped-wedge designs (where clusters are sequentially switched to treatment), and group sequential designs (where trials can be stopped early at prespecified interim analyses). Outcomes include risks, event rates, event-free time, and irregular visit-based scores, with follow-up times ranging from 28 days to 3 years after enrolment (Table 1). Consistent with real-world clinical trials, our simulated trials include participant dropout, missed visits, treatment switching, and site-level data errors that can only be detected by checking records against source documents.

For our main evaluation, we task the agent to act as a statistician working for the trial sponsor, the Ariadne Clinical Research Collaborative. The agent is provided with raw, messy trial data and trial documents, but not the analysis plan. It has to integrate the data from multiple linked source tables into an analysis dataset, with decisions needed for data versioning, participant identifiers, unit conversion, and resolving discrepancies between actual and recorded times. It then has to select an appropriate analysis method without knowing which standard assumptions are valid, and save the analysis code. Since we cannot evaluate the quality of the analysis code on a single trial (any estimate could be close to the truth by chance), we run the analysis code on 30 replication trials. These are new simulated trials with the same population, trial design, and sample size, but new participants and new data recording errors. The true effect is the same across all replication trials.

In TrialDev, an early choice determines what an agent can know, and earn, years later. We provide agents with observational data on 3 or 5 drug candidates, a budget, a 6-year time horizon, and a brief that includes the cost of each decision, the public drug approval criteria (significant benefit in confirmation trials and at least 300 treated participants completing the trial), and the dollar value of each unit of benefit and harm. We do not disclose what each drug does. Each drug candidate can be tested under standard or intensified dosing regimens, and the intensified regimen is treated as a distinct treatment option with its own benefit and harm trajectories over time. Agents must make strategic decisions about data acquisition. For each study, they need to determine the phase (early safety, exploration, or confirmation), which arms (treatment groups) to include, the number of participants, number of sites, follow-up duration, visit schedule, and when to view results. Agents can also make decisions about dropping arms or stopping a study early. Time costs money: participants only enrol as time progresses, and a study only reveals the data that have been paid for and observed. For a fair comparison, each drug development programme has a fixed participant pool where each participant has a complete trajectory under control and all drug candidates under both standard and intensified dosing regimens. This means that all agents draw from the same participant pool, and for any study an agent might purchase, there is a known answer. At any stage of the drug development programme, we can hand over the programme and replay it.

\begin{table}[t]
\centering
\caption{The two suites and their scope.}
\label{tab:glance}
\footnotesize
\setlength{\tabcolsep}{4pt}
\renewcommand{\arraystretch}{1.1}
\begin{tabular}{@{}>{\raggedright\arraybackslash\bfseries}p{0.13\linewidth}>{\raggedright\arraybackslash}p{0.405\linewidth}>{\raggedright\arraybackslash}p{0.405\linewidth}@{}}
\toprule
 & TrialEval: analyse a trial & TrialDev: develop a drug \\
\midrule
Public tasks & 420 tasks on 17 clinical questions, each at 5 context levels, from raw records to clean tables & 3,072 tasks on 384 programmes \\
Main evaluation & 84 tasks, raw records and no analysis plan, 24 models & 128 programmes at the generous budget, 10 models \\
Clinical scope & Childbirth safety to metastatic cancer, including heart failure, kidney disease and diabetes & Atopic dermatitis, chronic obstructive pulmonary disease, migraine and rheumatoid arthritis \\
Design & Individual and cluster randomisation, stepped-wedge rollout, group-sequential stopping, non-inferiority, treatment changes & 3 or 5 candidates, each at a standard and an intensified regimen; early safety, exploration and confirmation phases \\
Outcomes & Risks, event rates, time free of events, and scores at irregular visits & Benefit and harm over time, valued over the drug’s market lifetime \\
Size & 52 to 114,000 participants, at 3 to 400 sites & Frozen pool of 64,000 participants; 10--100, 30--1,500 or 300--3,000 per arm by phase \\
Time & Results from 28 days to 3 years after enrolment & 6 years to decide, with market exclusivity ending around year 18 \\
\bottomrule
\end{tabular}
\end{table}

Scoring against the truth changes observed capability. When we graded agents based on whether they used an accepted analysis method (an answer counted as correct only if it used a method from a predefined list and landed within tolerance of a target), 28\% of answers that were marked as incorrect were as accurate as a process-aware analysis that knows how each trial was set up. Opus 5.5 ranked 14th under this grading scheme, in contrast to the 2nd place ranking by its error against the true effect. Based on feedback from domain experts, we therefore fix the question for each task and evaluate any method by comparing against the true effect (as in simulation studies). We tested the simulator against this standard (Figure 2, Appendix C). When applied to simulated trials, the process-aware analysis recovers the specified effect (bias within 0.05 standard errors, with 95\% intervals containing the true effect 92--96\% of the time). Compared with registry and published trials held out from calibration, the size and conduct of our simulated trials are within or close to the range of real-world trials, and our effect sizes are similar to the distribution of effect sizes from Cochrane reviews. We also verify that two random halves of our tasks produce similar model scores (Spearman--Brown split-half reliability of 0.895), confirming that our scores can distinguish between models.

\begin{figure}[t]\centering
\includegraphics[width=1.0\linewidth,height=0.9\textheight,keepaspectratio]{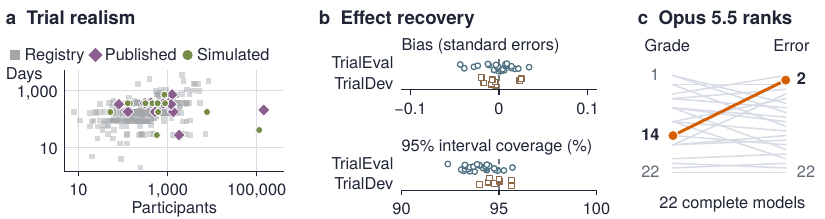}
\caption{Simulated trials are realistic, scores are reliable, and method-based grading misjudges agents. (a) Trial size and duration against registry trials and published examples. (b) Process-aware bias is near zero and 95\% interval coverage near 95\% in both suites. Points are configurations. (c) Accepted-method versus error ranks for 22 models. Opus 5.5 ranks 14th and 2nd. Of answers marked incorrect by method, 28\% match process-aware accuracy.}
\end{figure}

\section{Measures}

TrialEval error is the distance between an agent's estimate and the true effect, in units of the trial's noise. We use the standard error, which is how much an estimate would typically change if the same trial were run again on new participants. An effect and an error can be reported in the same unit but are different concepts: for example, if the true effect is −5 percentage points and the standard error is 2 percentage points, then the effect is 2.5 standard errors in size, whereas an analysis that on average produces estimates 3 points away from −5 has an error of 1.5. Even an unbiased analysis with normally distributed noise has an average error of 0.80, so a good score is close to 0.80 and not 0, whereas an agent that always guesses no effect has an error of 1.16.

We count a serious miss as occurring on this scale, i.e.~any error on a task that is at least 0.5 standard errors above that of the process-aware analysis.

We compare agents to 3 baselines, all run on clean analysis-ready data tables, and report agents as a percentage above the process-aware analysis. In actual trials, the analysis plan is typically locked in advance of seeing the data, which is not something that agents in our main evaluation have access to. For example, in a trial of a drug to prevent stroke, the primary analysis plan would typically be to compare stroke rates between the two trial arms based on site records (the standard analysis), and an alternative analysis plan would be to perform the same comparison after correcting the site records based on a randomly sampled audit against source documents. The plan's primary analysis (0.97) always follows the primary analysis in the plan, and always corrected (0.95) always performs the second analysis. The process-aware analysis (0.90) knows whether there are unequal recording errors between the two trial arms, but not what the true treatment effect is, and uses the arm-specific correction if there are unequal recording errors; otherwise it follows the plan.

TrialDev development value is a metric that unifies clinical benefit and harm, as well as money and time, into a single unit: value = clinical value − cost of harm − study costs − operating costs. The clinical value of an approved drug is the discounted profit of the drug, which is US\$250 million per year for every typical benefit unit (for migraine, this is 2 fewer migraine days per 28 days), from one year after the decision until loss of exclusivity around year 18, with a 10\% per year discount rate so that delays are penalised. The harm is converted to dollar value using a price specified in the brief, which is US\$12.5--20.8 million per percentage point of excess risk per year over the same period and depends on the disease. Operating costs are US\$15,000 per day. We compute the value assuming the true effects of each drug and report it as a percentage of the value that an oracle achieves on the same programmes. The oracle averages US\$343 million per programme, so each point is worth roughly US\$3.4 million. Stopping immediately leads to a value of 0\%, and spending without approval leads to a value below 0.

The oracle knows the true effects of every drug, but is restricted to searching a limited grid of development plans, locking in a plan, recruiting a finite number of participants for a trial, and going through the same statistical review process. It can therefore be unlucky (14 out of 104 of its advances fail the review process) and is a useful reference point rather than an upper bound. We also compare agents to a rule-based developer (RBD), which demonstrates how much value can be achieved through principled drug development using public information. It ranks drug candidates, sizes exploration and confirmation studies based on expected value, and only advances drugs if there is a significant result and the estimated profit is positive. It makes these decisions using a prespecified statistical model with a prior from Cochrane effect sizes.

\section{Results}

\begin{table}[t]
\centering
\footnotesize
\setlength{\tabcolsep}{2.5pt}
\caption{All models, ordered by average error on the main tasks, the score after accounting for task difficulty (Appendix B). Lower error and fewer serious misses are better; higher value is better. Value is development value as a share of the oracle's on the same programmes. Study cost is mean spending on studies relative to the RBD's. $^{\ddagger}$Open weight.}
\label{tab:leaderboard}
\resizebox{\linewidth}{!}{\begin{tabular}{@{}l c c c c c c c@{}}
\toprule
 & \multicolumn{3}{c}{TrialEval} & \multicolumn{4}{c}{TrialDev} \\ \cmidrule(lr){2-4}\cmidrule(lr){5-8}
Model & \begin{tabular}[b]{@{}c@{}}Average error\\(95\% interval)\end{tabular} & \begin{tabular}[b]{@{}c@{}}Serious\\misses, \%\end{tabular} & \begin{tabular}[b]{@{}c@{}}Interval\\coverage, \%\end{tabular} & \begin{tabular}[b]{@{}c@{}}Value, \% of oracle\\(95\% interval)\end{tabular} & \begin{tabular}[b]{@{}c@{}}Approved,\\\% of programmes\end{tabular} & \begin{tabular}[b]{@{}c@{}}Study cost,\\$\times$ RBD\end{tabular} & \begin{tabular}[b]{@{}c@{}}Finished,\\\%\end{tabular} \\
\midrule
Sonnet 5.5 & 0.93 (0.90--0.97) & 1 & 94 & 29 (14 to 40) & 45 & 1.4 & 100 \\
Opus 5.5 & 0.93 (0.90--0.97) & 1 & 95 & 48 (30 to 62) & 71 & 1.5 & 100 \\
Fable 5.1 & 0.94 (0.90--0.98) & 0 & 95 & -- & -- & -- & -- \\
Opus 5 & 0.95 (0.91--0.99) & 1 & 94 & -- & -- & -- & -- \\
GPT-6 Astra & 0.95 (0.93--0.98) & 4 & 95 & 14 ($-$1 to 26) & 45 & 2.5 & 100 \\
Sonnet 5 & 0.97 (0.94--1.00) & 5 & 94 & -- & -- & -- & -- \\
Opus 4.6 & 0.98 (0.95--1.03) & 6 & 94 & -- & -- & -- & -- \\
GPT-6 Sol & 0.99 (0.94--1.03) & 10 & 95 & 25 (0 to 44) & 47 & 2.3 & 100 \\
Qwen3.8 Max$^{\ddagger}$ & 0.99 (0.95--1.03) & 10 & 94 & $-$6 ($-$15 to 2) & 9 & 0.8 & 43 \\
Opus 4.8 & 0.99 (0.97--1.02) & 5 & 94 & -- & -- & -- & -- \\
Sonnet 4.6 & 1.00 (0.96--1.03) & 7 & 91 & -- & -- & -- & -- \\
GLM-5.3$^{\ddagger}$ & 1.00 (0.96--1.04) & 10 & 93 & 9 ($-$8 to 24) & 36 & 1.4 & 62 \\
GPT-5.6 Terra & 1.01 (0.97--1.07) & 8 & 95 & -- & -- & -- & -- \\
Kimi K3$^{\ddagger}$ & 1.02 (0.96--1.07) & 10 & 93 & 34 (18 to 46) & 51 & 1.6 & 95 \\
Mistral Large 4$^{\ddagger}$ & 1.04 (1.00--1.07) & 14 & 91 & -- & -- & -- & -- \\
DeepSeek V4.1 Flash$^{\ddagger}$ & 1.04 (1.01--1.08) & 13 & 92 & -- & -- & -- & -- \\
Opus 4.7 & 1.05 (1.01--1.08) & 14 & 91 & -- & -- & -- & -- \\
Gemini 4 Argon & 1.06 (0.99--1.15) & 13 & 93 & 2 ($-$11 to 13) & 19 & 1.1 & 84 \\
gpt-oss-120b$^{\ddagger}$ & 1.07 (1.01--1.13) & 15 & 89 & -- & -- & -- & -- \\
GPT-5.6 Luna & 1.08 (0.99--1.21) & 12 & 94 & -- & -- & -- & -- \\
GPT-5 & 1.11 (1.02--1.20) & 17 & 90 & -- & -- & -- & -- \\
Haiku 5.5 & 1.11 (1.04--1.19) & 21 & 92 & 11 ($-$1 to 22) & 29 & 0.8 & 100 \\
GPT-6 Luna & 1.18 (1.06--1.29) & 31 & 95 & $-$10 ($-$18 to $-$4) & 5 & 0.6 & 97 \\
DeepSeek V4 Pro$^{\ddagger}$ & 1.24 (1.15--1.35) & 26 & 93 & -- & -- & -- & -- \\
\midrule
Process-aware analysis & 0.90 (0.87--0.93) & 0 & 94 & -- & -- & -- & -- \\
Always corrected & 0.95 (0.92--0.98) & 0 & 95 & -- & -- & -- & -- \\
Plan's primary analysis & 0.97 (0.93--1.00) & 6 & 92 & -- & -- & -- & -- \\
Rule-based developer & -- & -- & -- & 58 & 54 & 1.0 & 100 \\
\bottomrule
\end{tabular}}
\end{table}

\subsection{The strongest models are competent analysts}

The strongest models are now able to perform the record reconciliation, data cleaning, and model fitting needed in trial analysis (Table 2, Figure 3). The earliest models we tested, released in August 2025, have 19--23\% higher errors than the process-aware analysis and have serious misses on 15--17\% of tasks. The top five models by error (Sonnet 5.5, Opus 5.5, Fable 5.1, Opus 5, GPT-6 Astra), all released since July 2026, now achieve errors of only 3.5--5.8\% above the process-aware analysis with 0--4\% serious misses, even when only given the raw data records. The true effect is captured within the 95\% confidence interval computed from the code they save 94--95\% of the time, which is similar to the process-aware analysis. They are also not swayed when an assumption is violated: their errors are insensitive to how much the standard analysis deviates from the truth, and the size of the true effect. The open-weight models still have 10--38\% higher errors than the process-aware analysis.

\begin{figure}[t]\centering
\includegraphics[width=0.9\linewidth,height=0.9\textheight,keepaspectratio]{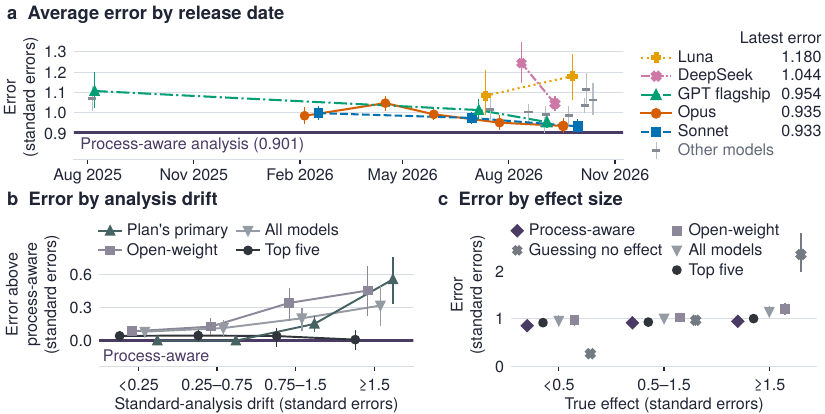}
\caption{Top models approach the process-aware analysis within about a year, and stay near it when the standard analysis drifts from the truth or true effects are large, unlike guessing no effect. (a) Average error of each model release by release date; lines join releases of one family and the key gives each family's latest error. GPT flagship comprises GPT-5, GPT-5.6 Terra and GPT-6 Astra, and Luna comprises GPT-5.6 Luna and GPT-6 Luna; grey ticks are other models; the purple line is the process-aware analysis. (b) Error above the process-aware analysis by how far the standard analysis lands from the true effect. (c) Error by size of the true effect. Errors and effects are in standard errors; vertical lines are 95\% bootstrap intervals.}
\end{figure}

Almost all of what separates the top five models from the process-aware analysis is a single decision: whether or not the evidence suggests that a correction is needed (Figure 4). The standard analysis is biased when the site records capture the outcome more poorly for one arm than the other due to unequal recording errors. A corrected analysis estimates the recording errors for each arm using the audited records. When the recording errors are equal, the standard analysis is sound and the correction only introduces noise. The top five models tend to check for discrepancies and then act as if they have performed a correction, regardless of whether it is needed: in tasks with unequal recording errors, their estimates are closer to the corrected analysis than the uncorrected one 99\% of the time, and 83\% of the time in tasks with equal recording errors. The same measure gives 100\% and 5\% (1 of 20 tasks) for the process-aware analysis, which uses the arm-specific correction only when recording errors are unequal; otherwise it follows the plan, which can itself draw on the audit. Because the cost of missing a necessary correction is much higher than performing an unnecessary correction, they match the performance of the process-aware analysis when a correction is beneficial and only incur a performance penalty when it is not (error of 1.15 vs 0.98). Only 3 models have better average errors than the always corrected analysis, so for most models a fixed rule that always corrects would be at least as accurate. While such a default strategy may be sufficient for an agent to get most of the way through an analysis, it is not sufficient to get through a drug programme, where each decision determines what evidence will be available for the next.

\begin{figure}[t]\centering
\includegraphics[width=0.9\linewidth,height=0.9\textheight,keepaspectratio]{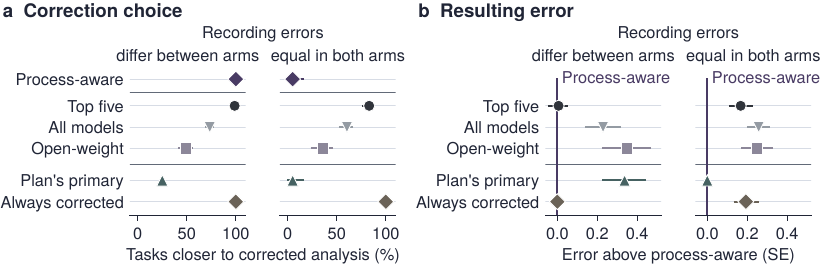}
\caption{The best models act as if they correct by default. (a) Fraction of tasks answered where the estimate is closer to the corrected than the uncorrected analysis, for tasks with different recording errors between arms and with the same recording errors. The process-aware analysis corrects when the errors are different (100\%) but rarely when the errors are the same (5\%); the top five models correct for both (99\% and 83\%). (b) Error compared to the process-aware analysis (vertical line) in standard errors. The top five match the process-aware analysis when the recording errors are different and incur 0.17 error when they are the same, which is close to always correcting (0.19); following the plan's primary analysis when the recording errors are different incurs 0.34 error. Error bars are 95\% bootstrap intervals.}
\end{figure}

\subsection{Equal analysts, unequal value}

Agents that are equally accurate analysts can return substantially different value (Figure 5). Opus 5.5, Sonnet 5.5, and GPT-6 Astra have indistinguishable analysis errors, but range from 14\% to 48\% of the oracle value, with Opus 5.5 outperforming GPT-6 Astra by 34 points (19 to 49). Better analysts tend to make better developers, with a negative rank correlation between error and value over the 10 agents (Spearman −0.64, 95\% CI −0.82 to −0.33). However, two agents may interpret a trial equally well and act differently on that interpretation.

\begin{figure}[t]\centering
\includegraphics[width=0.9\linewidth,height=0.9\textheight,keepaspectratio]{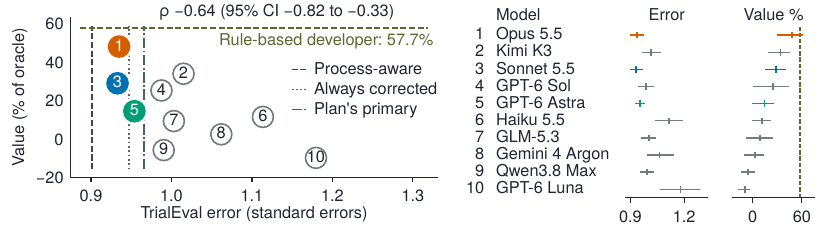}
\caption{Analysts with similar accuracy produce substantially different value. TrialEval average error against TrialDev value (\% of oracle) for ten agents, numbered by value from highest; filled points are the three lowest-error agents. The adjacent panels show each agent's error and value with 95\% intervals; dashed olive lines mark the rule-based developer's value. Spearman correlation is $-0.64$ (95\% CI $-0.82$ to $-0.33$), but Opus 5.5, Sonnet 5.5 and GPT-6 Astra have nearly identical errors and values from 14\% to 48\%.}
\end{figure}

\subsection{Where the value goes}

Most agents underperform the RBD, which achieves 57.7\% of the oracle value. The best agent, Opus 5.5, comes closest at 48\% (9.7 points lower, 95\% CI −6.3 to 26.9), while the remaining agents are 24 to 66 points worse than the RBD. On average, agents only use 43\% of their budget and 67\% of their six-year timeline, so money and time are not what holds most of them back.

Agents cast a wider net than the RBD but do not achieve better outcomes (Figure 6a). On programmes where the oracle advances a drug, agents test that drug 95\% of the time, confirm it 63\% of the time, and ultimately advance it 45\% of the time, whereas the RBD tests it 71\% of the time, confirms it 64\% of the time, and advances it 52\% of the time. Agents' 24-point advantage in testing the drug disappears by the time it reaches confirmation. On these programmes, agents stop at a similar rate to the RBD (30\% and 33\%), and the remaining gap is largely due to the 13\% of agent programmes that are left unfinished.

Most of the value gap is lost before the final decision (Figure 6b). Because every programme has an outcome, we can take the programme of an agent, pass it to the RBD after early safety or exploration, using the studies and costs so far, and replay the rest or just replace the final decision with a fixed rule. The differences between these replays assign the value gap to phases, with unfinished programmes counted separately. Of the average value gap of 40.8 points, 18.0 points are lost during exploration and 10.4 points are lost during confirmation; replacing the final decision alone recovers 0.5 points (−1.8 to 2.6).

Across agents, even if we use the true effects at the end, most of the value gap remains. If we fix the studies done by the agent and replace the final decision of the agent with a rule that advances the highest value treatment that passes review, this increases the value from 16.7\% of the oracle value to 17.2\% if we use the agent's estimates and to 23.2\% if we use the true effects, which is still much lower than the RBD value of 57.5\% on the same programmes.

\begin{figure}[t]\centering
\includegraphics[width=0.9\linewidth,height=0.9\textheight,keepaspectratio]{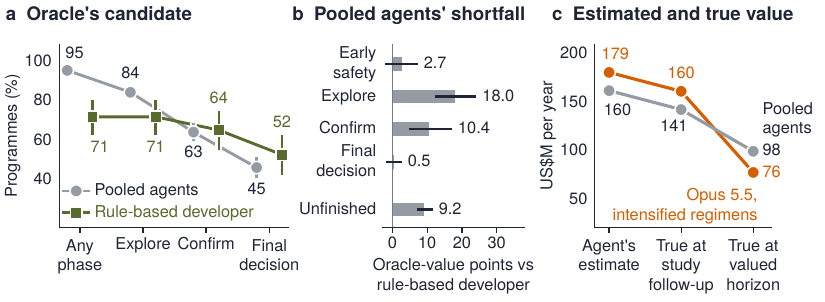}
\caption{Agents explore broadly without improving performance and most value is lost before the final decision. (a) Among programmes where the oracle advanced a candidate, the fraction that studied that candidate in any phase, explored it, confirmed it and advanced it at the final decision, for pooled agents and the rule-based developer. (b) Value loss of the pooled agents compared to the rule-based developer in points of oracle value by the stage of the value loss (Appendix E.1). (c) Annual net value of advanced treatments for Opus 5.5 with intensified regimens and for pooled agents: the agent's own estimate, then the true value at study follow-up and at the brief's valued horizon. Error bars in (a) and (b) are 95\% intervals.}
\end{figure}

Agents purchase more evidence than the RBD, but do not always purchase the evidence they need for their decisions. They spend around 40\% more on studies, yet 93 of the 165 failed approval reviews had fewer than 300 completed treated participants, which is below the public minimum stated in every brief. If we replay with waiting for enrolled participants to complete or increasing study size to meet this minimum, we gain 4.3 points (2.5 to 6.3), which is the largest gain out of all interventions we evaluated.

Stopping too early is a costly mistake, but the cost is driven by what agents stop, not how often they stop. Stopping on programmes where the oracle would advance a drug accounts for 15.6 points of the 40.8-point value gap, whereas programmes that end with the oracle's drug and regimen only account for 2.4 points. Agents that stop on programmes where the RBD would advance a drug lose US\$406 million per programme compared to the RBD, whereas agents that advance on programmes where the RBD would stop do not lose money on average, because a stop ends the drug's journey to patients, whereas a questionable advance still has to pass the approval review. If we treat unfinished programmes as not advanced, agents advance 59\% of programmes where the oracle has positive value and 32\% of programmes where it does not, compared to 76\% and 8\% for the RBD, so they are worse at knowing which drugs to develop.

Harm is the main additional cost for the top 3 agents (Table E.1), and shorter studies undercount it. Opus 5.5 creates 61 points more in clinical value than the RBD, but pays 62 points more in cost of harm by selecting the intensified regimen, which the RBD never tests. Agents do consider harm, but they underestimate how much benefit degrades and harm increases after a short study concludes, despite having access to longer follow-up. Intensified regimens behave differently over time, so an advance that looks profitable at the study's follow-up can lose value by the valued horizon, the time point set in the brief at which benefit and harm are valued (Appendix E.3). For the intensified regimens that Opus 5.5 advances, it estimates a net value of US\$179 million per year, whereas the true net value is US\$160 million at the study's follow-up and US\$76 million at the brief's valued horizon (Figure 6c). Its reading of the study is close; its extrapolation beyond it is not. Of its 91 approved drugs, 17 (19\%) lost money at the true effects, which cost 9.8 points, as large as its entire gap to the RBD; it predicted a profit for 16 of those, so its own estimates gave no warning of these losses.

\section{Discussion}

In just over a year, the strongest agents have transitioned from frequent, large-scale errors to near parity with a process-aware analysis while reading raw records, even when an assumption is violated. Agents have become competent analysts, but they are not yet reliable long-horizon decision-makers. Agents that read trials equally well can make decisions of substantially different value, with most value lost far in advance of the final decision through study acquisition, abandonment of promising drugs, or misplaced confidence in short trials. Benchmarks which compare against recorded answers cannot determine where this value is lost as there is no outcome for the paths an agent did not take.

For practitioners interested in agent deployment, human oversight should be provided where value is lost: study design, harm valuation at the relevant time horizon, and decision review to prevent premature development stoppage. For researchers, the simulator provides a potential future training environment as the true value of each decision is known. Natural extensions are to give agents knowledge of prior trials in the same disease area, and to run programmes in sequence to test whether agents learn from earlier outcomes. We release our benchmark, drug development environment, replication trials, the rule-based developer, and independent verifier (https://github.com/BradSegal/TrialAgentBench) to enable the community to determine if it is possible to bridge the gap between capable analyst and reliable decision-maker through tool use, training, or oversight. For agents to get better medicines to patients sooner, they must not only analyse but design studies with the subsequent decision in mind: whether the benefit lasts, whether the harm outweighs it, and whether the drug is promising enough to warrant further testing.

\begin{ack}
We thank Dan Manela and Joshua Fieggen from the University of Oxford for their feedback and insight on trials and statistics.

Bradley Segal was funded by a Rhodes Scholarship. David Clifton was funded by an NIHR Research Professorship (NIHR302440), a Royal Academy of Engineering Research Chair, and the InnoHK Hong Kong Centre for Cerebro-cardiovascular Engineering (COCHE), and was supported by the National Institute for Health Research (NIHR) Oxford Biomedical Research Centre (BRC) and the Pandemic Sciences Institute at the University of Oxford. William Bolton and Bradley Segal acknowledge funding from ARIA and DSIT and Pillar VC under the Encode: AI for Science Fellowship. The authors declare no competing interests.
\end{ack}

\clearpage
\bibliographystyle{unsrtnat}
\bibliography{references}
\clearpage
\appendix
\makeatletter
\setlength{\@fptop}{0pt}
\makeatother
\FloatBarrier
\section{The environments}
\setcounter{table}{0}\setcounter{figure}{0}\renewcommand{\thetable}{\Alph{section}.\arabic{table}}\renewcommand{\thefigure}{\Alph{section}.\arabic{figure}}

TrialEval and TrialDev share the same clinical trial simulation compiler but differ in the agent's objective. In TrialEval, the agent must read a single completed trial, whereas in TrialDev, the agent must purchase evidence across a development programme.

\begin{table}[!htbp]
\caption{Terms used throughout the appendix.}
\centering\scriptsize\setlength{\tabcolsep}{2pt}
\begin{tabular}{>{\raggedright\arraybackslash}p{0.199\linewidth}>{\raggedright\arraybackslash}p{0.757\linewidth}}
\toprule
\textbf{Term} & \textbf{Meaning} \\
\midrule
World & One simulated setting that fixes the population, the true effect, the trial design, and its size; any number of trials can be drawn from it \\
True effect & The prespecified comparison of complete outcomes with every simulated participant on treatment versus control; known without an estimator \\
Replication trial & A new simulated trial from the same world: new participants and new recording errors, the same true effect \\
Standard error & How much an estimate would typically change if the same trial were rerun with new participants; errors and effects are reported in these units \\
Error & Absolute distance between an estimate and the true effect, in standard errors of a trial of that size; average error adjusts for task difficulty \\
Serious miss & A task on which an agent's error is at least half a standard error above the process-aware analysis \\
Unequal errors & Routine records miss more true outcomes in one arm, and a standard analysis is biased \\
Equal errors & Routine records miss outcomes at the same rate in both arms, and a standard analysis is valid \\
Standard analysis & The analysis plan's default route, which assumes recording error is equal across arms \\
Correction & An audit-based analysis that estimates recording error arm by arm and pays for it in variance \\
Process-aware analysis & Knows whether recording errors are equal and uses the standard analysis or the correction accordingly \\
Model, agent & A model is the AI system; an agent is that model running in the benchmark harness \\
Main tasks, main evaluation & The 84 TrialEval tasks at context level C5 and the 128 TrialDev programmes at the generous budget \\
Candidate, regimen & A drug in a programme, at a standard or intensified dose \\
Intensified regimen & A candidate at its intensified dose: a separate treatment option with its own benefit and harm over time \\
Advance & The agent's final decision to put one candidate at one regimen to review \\
Approved & A candidate that passes review: significant benefit in the confirmation trials and at least 300 completed treated participants \\
Stop, unfinished & Ending a programme without advancing; a programme that reaches a run limit before any final decision \\
Value & Discounted profit from an approved drug at the true effects, minus the cost of harm, study costs and daily operating costs \\
Clinical value & The profit term of value, from true benefit, before any cost \\
Typical benefit unit & The typical effect size for the disease; it earns US\$250 million a year \\
Cost of harm & The price in the brief for true excess risk, from launch to the end of exclusivity \\
Point & One percentage point of oracle value on the same programmes \\
Oracle & Knows every drug's effects but searches a limited grid of plans and faces the same statistical review; a reference point, not an upper bound \\
Rule-based developer & Takes seven decisions in a fixed order, using a conventional statistical model fitted in advance on separate simulated development programmes and then frozen. Abbreviated RBD \\
\bottomrule
\end{tabular}
\end{table}

\subsection{A clinical trial simulation compiler}

Our simulation compiler takes a specification of a clinical trial and produces a programmable simulated clinical trial whose truth is known. The specification includes a clinical question, a population, a trial design, trial conduct, and a data capture process. A copula-based model samples correlated features for participants, a treatment assignment and disease progression process assigns outcomes to participants under each treatment, and a data capture process produces the data a trial analyst would receive. The trial data are stored as the documents and datasets of a drug sponsor.

Because our simulation compiler is programmable, the trials can differ substantially. They can randomise individuals or clusters, roll out treatment in a staggered fashion, or stop early at a prespecified interim analysis. They study a range of outcomes including risks, time-to-event, event rates, and irregularly measured scores, and range from 52 to 114,000 participants. They include realistic data issues such as participants discontinuing, switching treatments, missing visits, or dropping out, and routine data capture may include errors that can be detected by random audits.

A Gaussian copula model \cite{sklar1959fonctions, nelsen2006copulas} separates the correlation structure of the features from their marginal distributions. For each clinical question, we specify a target rank correlation \(\rho_{jk}\) between features \(j\) and \(k\) and a clinical marginal distribution \(F_j\) for each feature. We sample a latent Gaussian vector \(Z\) with a correlation matrix \(R\) tailored to achieve the desired rank correlation. We then transform each latent value to a value \(V_j\) in the unit interval using the Gaussian cumulative distribution function \(\Phi\). Finally, we map each \(V_j\) to the clinical scale using the inverse cumulative distribution function of \(F_j\). Because both transformations are monotonic, they preserve the rank of each participant.

\[Z \sim \mathcal{N}(0, R),\quad R_{jk} = 2\sin\!\left(\frac{\pi}{6}\rho_{jk}\right),\qquad V_j = \Phi(Z_j),\qquad X_j = F_j^{-1}(V_j).\]

Given these features, we use a structural causal model \cite{pearl2009causality} to map participant features to treatment assignment, disease progression, clinical outcomes, and data capture up to a data cut-off date.

\[\begin{gathered}
A = g_A(X, U_A; \text{design}),\\
H(t+1) = g_H(H(t), X, A(t), U_H(t)),\qquad Y(t) = g_Y(H(t), X, A(t), U_Y(t)).
\end{gathered}\]

Treatment \(A\) is determined by the trial design. Disease state \(H\) evolves in discrete time steps based on the previous disease state, participant features, and treatment. The clinical outcome \(Y\) is a function of the disease state. We run the last two equations twice for each participant, once with \(A\) set to treatment and once with \(A\) set to control, keeping the same participant features and noise terms. This allows us to compute the potential outcomes \cite{imbens2015causal} for each participant, and the true treatment effect is a prespecified comparison of the potential outcomes across the population. Finally, the data capture process determines which participants attend study visits, which are lost to follow-up, and which outcomes are recorded with errors.

We calibrate the parameters of the simulated clinical trials to published data. We set the trial size to the median number of randomised participants in matching trials from the AACT database \cite{tasneem2012aact}. If there are not enough matching trials, we use the published trial size or power analysis from the study, and we use a larger size where a smaller trial could not identify the effect. We set the error rate of the data captures, and the difference in error rates by arm, based on published comparisons between site or routine data capture and central or adjudicated data capture \cite{goldmacher2026setting, tang2010influence, goonasekera2024accuracy}. We set the true treatment effects by sampling from a mixture model fit to 23,551 studies from the Cochrane database \cite{vanzwet2021statistical}, and keep only effects that are possible for each outcome; for example, a risk reduction cannot exceed the control risk.

\begin{figure}[!htbp]\centering
\includegraphics[width=1.0\linewidth,height=0.9\textheight,keepaspectratio]{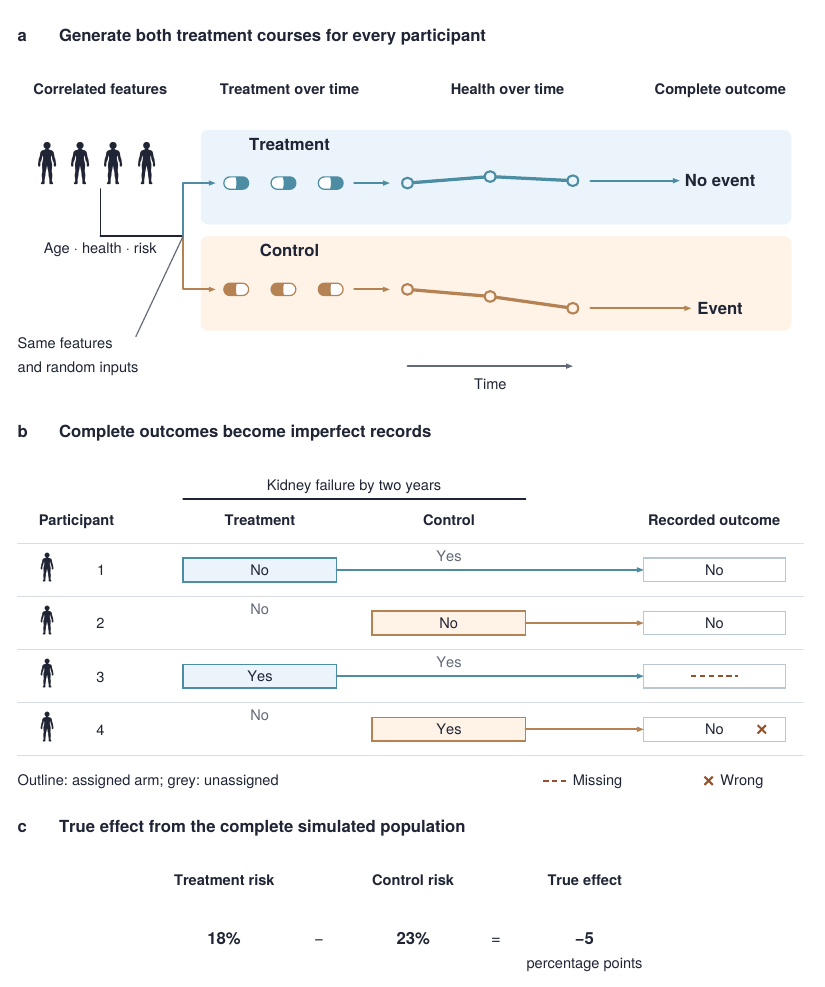}
\caption{The simulator creates the full outcome and then the trial records, so the true effect is known. (a) For a given set of participant features and random numbers, the full clinical trajectory and outcome is produced for both treatment and control before an arm is assigned. (b) For each participant, the assigned arm selects which outcome is used, and the trial record may have missing or incorrect entries. (c) The true effect is computed from the full population. Here the treatment risk is 18\% and the control risk is 23\%, so the true effect is −5 percentage points.}
\end{figure}

\begin{figure}[!htbp]\centering
\includegraphics[width=1.0\linewidth,height=0.9\textheight,keepaspectratio]{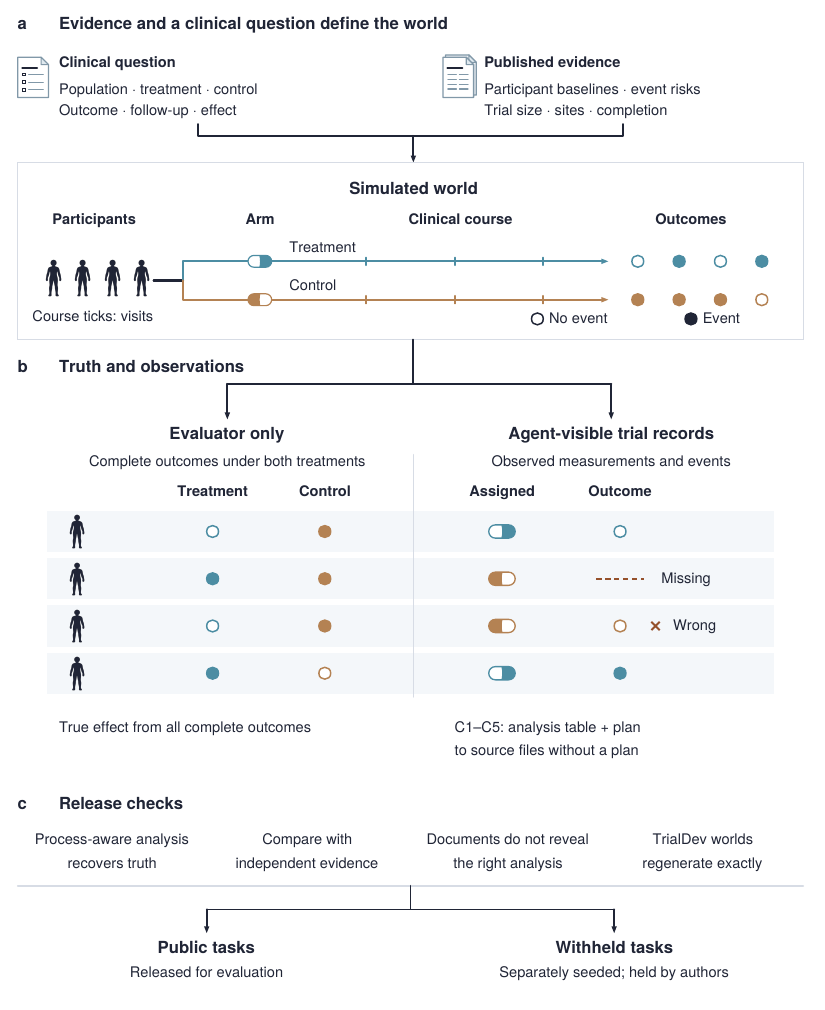}
\caption{Each world is fixed before any trial records exist, and the agent only sees the trial records. (a) A clinical question and published evidence define the world. (b) The evaluator has access to the complete outcomes and true effect; the agent only sees trial records at context levels C1 to C5 (Appendix A.2). (c) Pre-release checks test that the process-aware analysis recovers the true effect, trials compare with independent evidence, documents do not reveal the correct analysis, and TrialDev worlds can be exactly regenerated.}
\end{figure}

\subsection{TrialEval tasks}

In a TrialEval task, the agent must act as the trial statistician. It must read the clinical trial documents, reconstruct the analysis datasets from the raw trial data, and provide an answer to the clinical question in the form of a point estimate and an interval. The agent has an offline sandbox where it can run code in Python or R, with at most 50 turns and 1 hour for the main tasks. The agent's analysis is saved as a code script, which is rerun before the final submission. We evaluate the agent's code script on replication trials (Appendix B).

TrialEval includes 17 clinical questions and 21 scenarios, where a scenario is a particular combination of a clinical question and a data capture pattern. Each clinical question has 4 worlds with different true effects, resulting in 68 base worlds and 84 main tasks. The main tasks derived from the same base world are not statistically independent, so we take care to keep the tasks together when creating data splits and bootstrapping datasets. Each task has 5 different context levels, resulting in 420 public tasks. We also reserve 68 tasks with unique base worlds, some of them from scenarios exclusive to the reserved split.

\begin{table}[!htbp]
\caption{The 17 clinical questions cover a wide range of study designs, summary measures, and trial sizes ranging from 52 to 114,000 randomised participants.}
\centering\scriptsize\setlength{\tabcolsep}{2pt}
\begin{tabular}{>{\raggedright\arraybackslash}p{0.351\linewidth}>{\raggedright\arraybackslash}p{0.216\linewidth}>{\raggedright\arraybackslash}p{0.210\linewidth}>{\raggedright\arraybackslash}p{0.156\linewidth}}
\toprule
\textbf{Clinical question} & \textbf{Trial design} & \textbf{Summary measure and time point} & \textbf{Randomised} \\
\midrule
Childbirth in maternity units: postpartum safety pathway vs usual care, on severe haemorrhage, laparotomy for bleeding or death from bleeding & Stepped-wedge cluster-randomised & Risk difference by day 42 & 114,000 \\
Chronic heart failure after hospitalisation: remote management vs usual care, on the number of heart-failure hospitalisations & Parallel group & Rate ratio through day 365 & 320 \\
Chronic heart failure: device-guided vs standard care, on heart-failure hospitalisation & Parallel group & Risk difference by day 548 & 676 \\
Chronic kidney disease: intensive management vs usual care, on kidney failure or death & Parallel group, with a fixed testing sequence & Difference in restricted mean event-free time through day 1095 & 580 \\
Chronic kidney disease: kidney-protection therapy vs usual care, on albuminuria & Parallel group, repeated measurements at irregular visits & Mean difference at day 365 & 128 \\
Long-term-care residents: vaccination vs usual programme, on laboratory-confirmed respiratory infection & Parallel cluster-randomised & Risk difference by day 180 & 7,680 \\
Metastatic colorectal cancer, first line: combination vs standard therapy, on objective tumour response & Parallel group & Risk difference by day 168 & 660 \\
Metastatic colorectal cancer: maintenance therapy vs observation, on death from any cause & Parallel group & Risk difference by day 730 & 492 \\
Outpatient COVID-19: antiviral regimen vs supportive care, on sustained clinical recovery & Parallel group & Risk difference by day 28 & 576 \\
Parkinson disease: rehabilitation vs usual care, on a motor-disability score & Parallel group & Mean difference at day 180 & 52 \\
Relapsing multiple sclerosis: disease-modifying vs standard therapy, on confirmed disability progression & Parallel group, non-inferiority, with treatment changes recorded & Risk difference by day 730 & 852 \\
Severe asthma: add-on biologic vs placebo, on a first severe exacerbation & Parallel group, with treatment changes recorded & Risk difference by day 365 & 600 \\
Severe uncontrolled asthma: add-on biologic vs placebo, on the number of severe exacerbations & Parallel group & Rate ratio through day 365 & 880 \\
Type 2 diabetes: add-on glucose-lowering therapy vs placebo, on glycated haemoglobin & Parallel group & Mean difference at day 365 & 456 \\
Type 2 diabetes: metabolic therapy vs placebo, on major cardiovascular events & Parallel group, standardised to participants with BMI at least 35 & Risk difference by day 1095 & 4,792 \\
Uncontrolled persistent asthma: add-on inhaled therapy vs usual care, on an asthma-control score & Parallel group, repeated measurements & Mean difference at day 365 & 144 \\
Unprovoked venous thromboembolism: extended vs standard-duration anticoagulation, on recurrence & Group-sequential parallel group & Risk difference by day 180 & 616 planned; enrolment can stop earlier \\
\bottomrule
\end{tabular}
\end{table}

Twelve of the scenarios (48 tasks) have standard data capture, and can be correctly analysed under standard assumptions. The other 9 scenarios have a random subset of the data audited against source documents. Only the data capture process differs from the standard scenario: 4 scenarios (16 tasks) have unequal recording errors and 5 scenarios (20 tasks) have equal recording errors. For scenarios with unequal errors in the capture of clinical events, we set the ratio of odds ratios between site data and audit data to be within the published confidence interval for unblinded site assessors against central assessors \cite{ndoungadiakou2016comparison}.

\begin{figure}[!htbp]\centering
\includegraphics[width=1.0\linewidth,height=0.9\textheight,keepaspectratio]{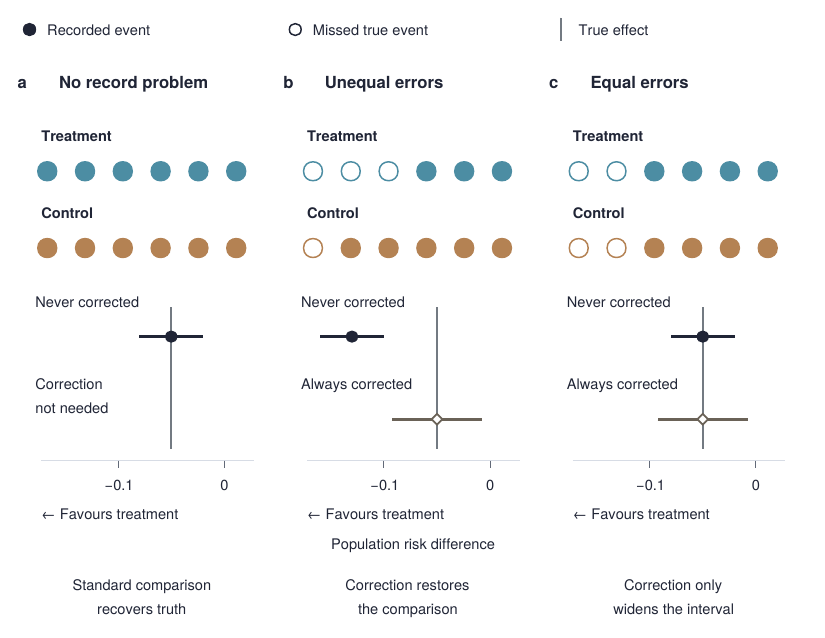}
\caption{Recording errors bias the standard treatment comparison only when they differ between arms. Filled circles are recorded events and open circles are true events missing from the records; below, never-corrected (dot) and always-corrected (diamond) estimates with confidence intervals against the true effect (vertical line), on a schematic risk-difference axis. (a) With complete records, the standard comparison recovers the true effect. (b) With unequal errors, it is displaced and correction restores it. (c) With equal errors, it remains valid and correction only widens the interval.}
\end{figure}

In context levels C1 and C2, the data provided to the agent is in the form of analysis-ready tables in standard formats such as SDTM and ADaM. For example, the data may be provided as a table with one row per participant and a separate table for time-to-event data. In context levels C3 and C4, the agent must reconstruct the analysis-ready tables from the raw source data tables. In context level C5, which includes all 84 main tasks, the agent must also deal with data corrections, unit errors, and time misalignments in the raw source data tables, and is not given an analysis plan.

\begin{figure}[!htbp]\centering
\includegraphics[width=1.0\linewidth,height=0.84\textheight,keepaspectratio]{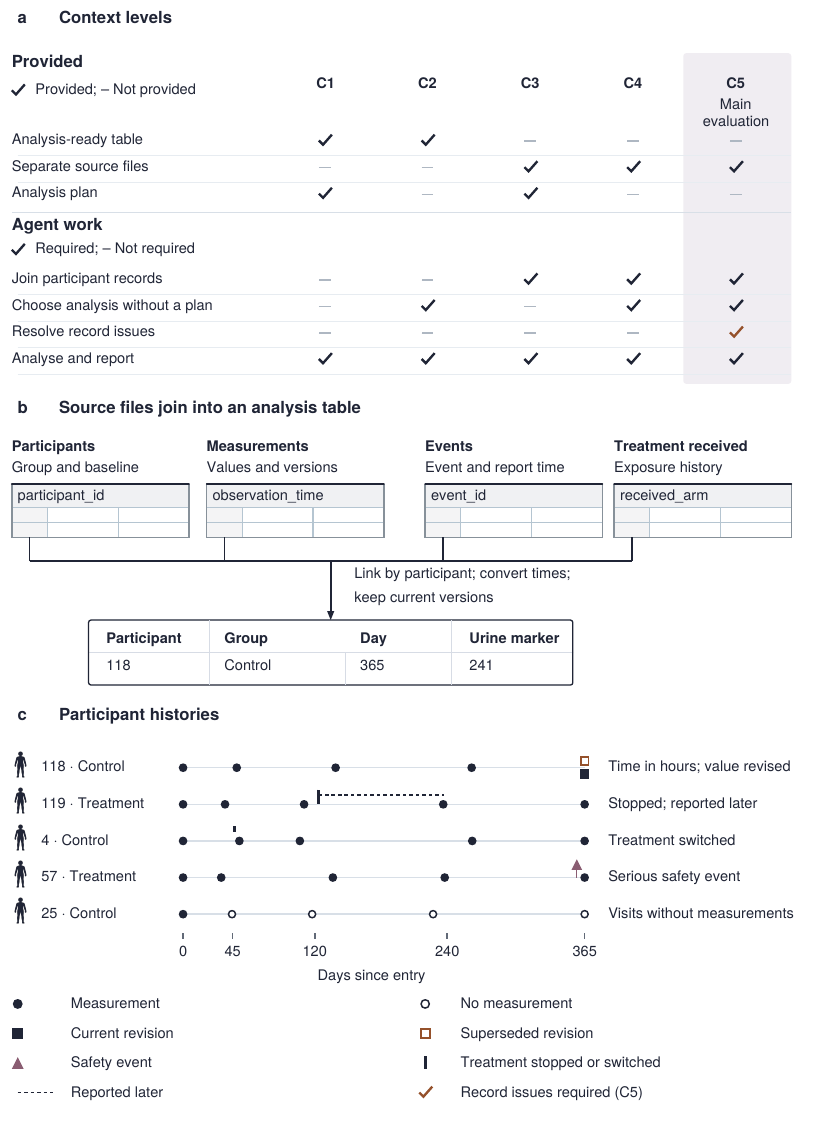}
\caption{Context levels vary in what they provide versus what the agent must determine. (a) Each column represents a different context level; a tick indicates what is provided or what the agent needs to do. (b) The agent links source files by participant, converts times, and retains the latest versions to create a unified analysis table. (c) Each row represents a participant's history over days since entry. C1 supplies analysis tables and a locked plan; C2 requires analysis choice. C3 requires reconstruction of source tables with a locked plan; C4 also requires analysis choice. C5 adds record reconciliation.}
\end{figure}

A C5 task is a substantial data curation job. Each task consists of 10 to 13 files, which include 3 to 5 source data tables (linked together) and 7 to 8 documents. The combined source data tables have between 48 and 67 columns (median of 56) and between 311 and 888,520 rows of data (median of 4,489). The agent must reconstruct the trajectory of each participant from data tables with different participant identifiers, deal with data versioning, data retransmission, and unit shifts, and align different notions of time (event time, observation time, documentation time, and data availability time) with the clinical question.

\begin{figure}[!htbp]\centering
\includegraphics[width=1.0\linewidth,height=0.9\textheight,keepaspectratio]{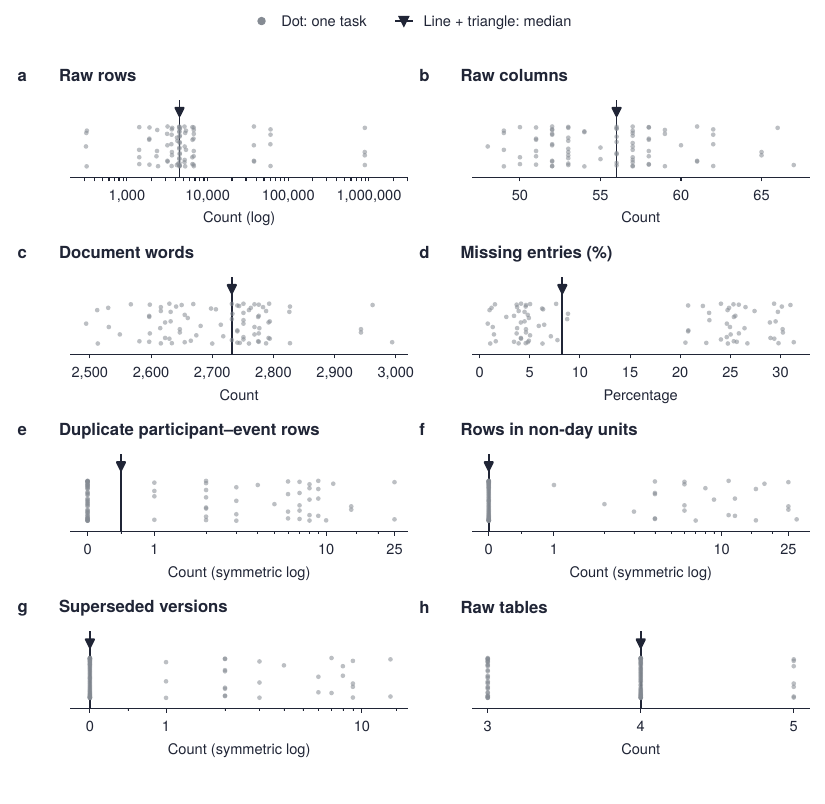}
\caption{Main tasks exhibit wide variability in trial record reconstruction requirements, though most have minimal defective rows. Each point represents one of the 84 main tasks, with the vertical line indicating the median. Panels show raw row count, raw column count, document word count, missing entries (\%), duplicate participant--event rows, rows in non-day units, superseded versions, and raw table count.}
\end{figure}

\begin{figure}[!htbp]\centering
\includegraphics[width=1.0\linewidth,height=0.9\textheight,keepaspectratio]{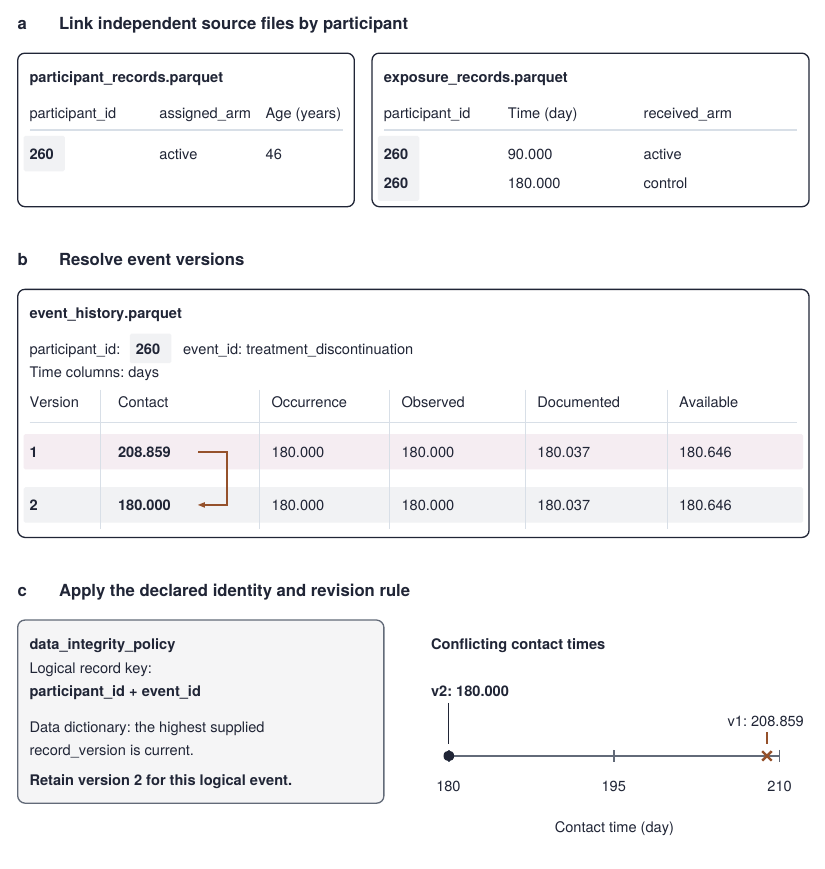}
\caption{Obtaining a single analysis value requires multiple reconciliation steps, illustrated on raw source records from one main task. (a) Participant identifiers are used to link different source files. (b) Two versions of a treatment discontinuation event show discrepancy in contact time (208.859 vs 180.000). (c) The data integrity policy specifies that the highest provided version is considered current, so the agent retains version 2.}
\end{figure}

\subsection{The frozen participant pool}

In a TrialDev programme, the agent plays the role of a drug developer. It is given observational data for 3 or 5 drug candidates, a budget, a time limit of 6 years, and a brief that includes the cost structure and approval rule but no details on the drug properties. The agent makes decisions on purchasing studies across three phases and observes the outcomes of these studies over time. The agent either decides to stop or advance a single drug candidate at a single regimen, or it reaches a limit of 100 turns. The agent's actions are evaluated according to the value they generate (Appendix B).

We simulate a base population of 64,000 participants with the full progression of their disease over time under control and under each drug candidate at both regimens before any agent interacts with the environment. Thus, for 3 drug candidates, there are 7 copies of the participants and for 5 drug candidates, there are 11 copies, giving a total of 448,000 or 704,000 participant trajectories per programme. All studies that the agent can purchase have pre-determined answers.

\begin{figure}[!htbp]\centering
\includegraphics[width=1.0\linewidth,height=0.87\textheight,keepaspectratio]{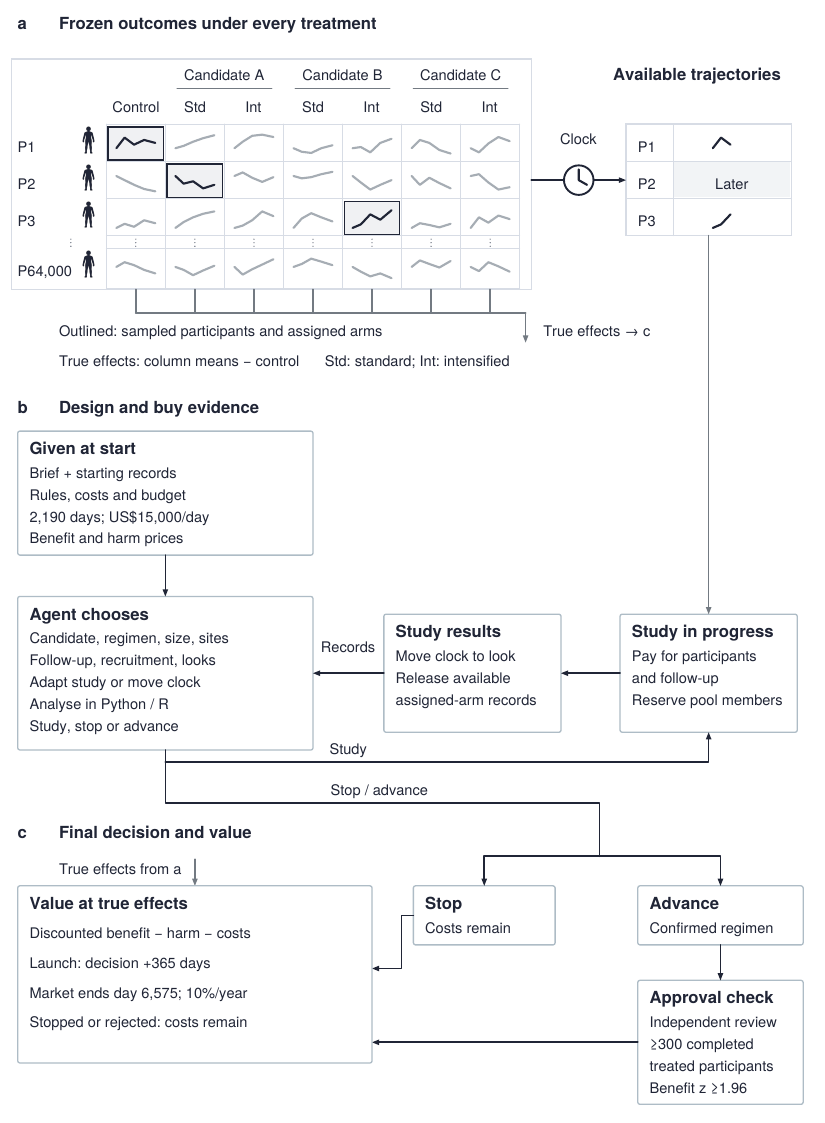}
\caption{Studies reveal assigned-arm records from outcomes fixed before the programme starts. (a) Rows P1, P2, P3 and P64,000 illustrate the first, second, third and last participants in the frozen pool of 64,000. Outlines identify sampled participants and assigned arms. True effects are column means minus control at final follow-up. Std and Int mean standard and intensified regimens. (b) The agent receives the brief and starting records, buys studies and updates its choices after results. (c) It stops or advances one confirmed regimen for approval review.}
\end{figure}

When the agent purchases a study, it views a subset of the participant pool. Once the agent commits to a study design, participants are sampled based on the sites that are activated, the screening criteria, and the recruitment strategy. The participants are then randomised and scheduled to the study, and are only enrolled once the programme time reaches the scheduled time. A results release is a pre-determined snapshot of the data at a data cut-off. It only reveals the treatment assignment and the data that has been collected up to that point, which includes participant dropout, missed visits, and noisy measurements. Participants enrolled in one study are not re-used for other studies. The random seed used for participant recruitment is fixed for each programme, so if two drug developers make the same study design decisions in the same order, they will observe the same study results.

\subsection{Studies, costs and the clock}

Evidence is not free: it incurs both monetary and time costs, with time being the most significant cost. To purchase a study, the agent must specify the following design elements and pay the associated costs:

\begin{table}[!htbp]
\caption{Each study provides numerous choices for the agent, constrained by public limits.}
\centering\scriptsize\setlength{\tabcolsep}{2pt}
\begin{tabular}{>{\raggedright\arraybackslash}p{0.110\linewidth}>{\raggedright\arraybackslash}p{0.538\linewidth}>{\raggedright\arraybackslash}p{0.296\linewidth}}
\toprule
\textbf{Choice} & \textbf{What it means} & \textbf{Limit} \\
\midrule
Phase & The stage of development. Early safety is a small study of harm, exploration compares candidates and regimens, and confirmation is the larger study that the approval review uses & Each candidate and regimen needs study results from the phase before \\
Arms & Which candidate and regimen pairs to test against a shared control group & 1--6 pairs, each sized separately \\
Size per arm & How many participants to enrol in each arm & Early safety 10--100, exploration 30--1,500, confirmation 300--3,000; at most 8,000 per study and 16,000 per programme \\
Sites & How many hospitals or clinics recruit participants & From 2 up to the whole network, which has 88, 115, 139 or 338 sites depending on the programme \\
Follow-up & How long each participant is observed & A shorter and a longer option per phase \\
Schedules & When participants attend assessment and safety visits. Fewer visits cost less and show less & Any days from the brief's visit grid up to the follow-up; in migraine, days 28, 56, 84, 168 and 336 after enrolment \\
Results releases & When the agent sees results: on a programme day it sets, or earlier once a number of participants it sets have finished follow-up & 1 or 2 per study; each uses up one planned look, even if few participants have finished \\
Recruitment & Which screened people enrol. By default every eligible person does. The agent can accept fewer, overall or within groups the brief names, such as participants with a prior severe event in COPD, to focus a study on those it expects to benefit most. Screening is charged even if the person does not enrol & 10--100\% overall and per group \\
Interim pause & Enrol a first group, see its results, then resume recruitment or close the study, so the agent need not commit the full cost up front & Optional \\
During a study & What can change after a study starts & Close it, resume recruitment, drop an arm or reduce the control group \\
Before committing & A preview of the design and its costs & Commits nothing \\
\bottomrule
\end{tabular}
\end{table}

\begin{table}[!htbp]
\caption{Setup, site, and screening costs increase with each phase. Each cell shows the cost per unit or the number of days for study setup and closeout; costs are identical across all programmes and profiles.}
\centering\scriptsize\setlength{\tabcolsep}{2pt}
\begin{tabular}{>{\raggedright\arraybackslash}p{0.134\linewidth}>{\raggedright\arraybackslash}p{0.073\linewidth}>{\raggedright\arraybackslash}p{0.110\linewidth}>{\raggedright\arraybackslash}p{0.086\linewidth}>{\raggedright\arraybackslash}p{0.098\linewidth}>{\raggedright\arraybackslash}p{0.171\linewidth}>{\raggedright\arraybackslash}p{0.086\linewidth}>{\raggedright\arraybackslash}p{0.127\linewidth}}
\toprule
\textbf{Phase} & \textbf{Set-up} & \textbf{Per site} & \textbf{Per screened} & \textbf{Per enrolled} & \textbf{Per participant-day} & \textbf{Per visit} & \textbf{Start-up, close-out (days)} \\
\midrule
Early safety & US\$1.0M & US\$50,000 & US\$1,000 & US\$12,000 & US\$60 & US\$1,500 & 60, 14 \\
Exploration & US\$2.0M & US\$75,000 & US\$1,500 & US\$10,000 & US\$40 & US\$600 & 90, 30 \\
Confirmation & US\$4.0M & US\$100,000 & US\$2,000 & US\$12,000 & US\$40 & US\$800 & 120, 45 \\
\bottomrule
\end{tabular}
\end{table}

The agent has 6 core actions it can take: create a study (previewing the design, cost, or committing), request a results release, adapt a study, wait, stop, and advance. If the agent attempts to design a study outside of the design limits it will receive a message indicating the design limit violated and the acceptable range. The design limits only prevent the agent from designing impossible or unaffordable studies; it is still possible for the agent to purchase a study that is too small to meet the approval rule (which is public information included in the brief and only applied at the final review).

Enrolling a participant creates a study liability: the cost of all the remaining follow-up days and scheduled visits for that participant. For example, if an agent has recruited 40 participants, then decides to stop enrolment, it will not incur any additional enrolment costs. However, it will still be charged for the remaining visits of those 40 participants. The total amount of money spent on studies plus the total study liability must not exceed the budget.

Time within the programme is a shared resource across all studies. Time is limited to a maximum of 2,190 days (6 years) and only advances when the agent requests results, waits, or makes a final decision. An approved drug generates profit until its exclusivity period ends on day 6,575 (about year 18).

\subsection{The size of the decision space}

If we only consider one balanced study per phase and a final decision, there are approximately \(10^{21}\) to \(10^{27}\) combinations of decisions, depending on the disease area and the number of candidates. For example, consider the rheumatoid arthritis programme with 3 candidates. There are 3 candidates at 2 regimens, for a total of 6 active arms. Any non-empty subset of the 6 active arms can be compared to a common control arm, so there are \(2^6 - 1 = 63\) possible sets of arms. A balanced study has one size per arm, including the control arm: 10 to 100 participants for early safety (91 values), 30 to 1,500 for exploration (1,471 values), and 300 to 3,000 for confirmation (2,701 values). There are 139 sites in the network, and the agent can open from 2 to 139 sites (138 values). There are 2 follow-up periods for each phase. This means that for a confirmation study alone, there are 63 × 2,701 × 138 × 2 = 46,964,988 designs. Multiplying the three phases and the 7 final decisions (stop or advance one of the 6 treatments), we get approximately 1.3 × \(10^{22}\) combinations of decisions. This count includes some combinations that the participant limits would rule out, and does not include unbalanced arms, repeated studies, second results releases, visit schedules, recruitment shares, or study adaptations.

The RBD (Appendix B) only searches a small fraction of this decision space. Prior to making its first decision, it evaluates the number of top candidates to test (up to 3 or 5), 12 different exploration study sizes, 6 different site counts, and 96 different confirmation study designs (2 follow-up periods × 8 sizes × 6 site counts): 3 × 12 × 6 × 96 = 20,736 plans for 3 candidates and 5 × 12 × 6 × 96 = 34,560 plans for 5 candidates.

Agents do take advantage of the decision space: 27\% of attempted study designs use the maximum of 6 active arms, 28\% plan for 2 results releases, and 47\% set non-standard visit schedules. On one programme, the oracle, the RBD, and Opus 5.5 all advance the same candidate, but Opus 5.5 tests more active arms and advances the intensified regimen.

\begin{figure}[!htbp]\centering
\includegraphics[width=1.0\linewidth,height=0.83\textheight,keepaspectratio]{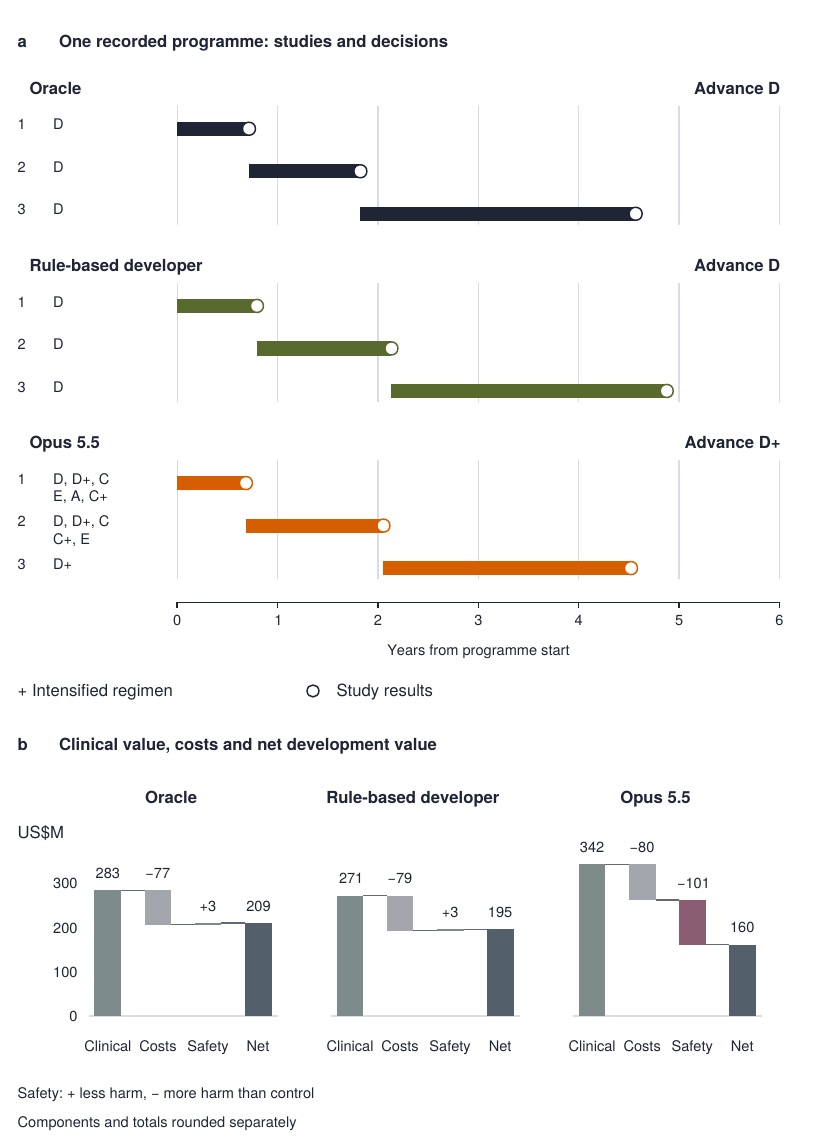}
\caption{On one programme with 5 candidates and 2 regimens each, Opus 5.5 achieves the highest clinical value but the lowest net value due to the cost of harm. (a) Each row represents a study; a bar indicates its duration from start to results, an open circle marks the results release, and a plus sign indicates the intensified regimen. The oracle and rule-based developer advance candidate D at the standard regimen; Opus 5.5 advances the intensified regimen D+. (b) Clinical value (US\$283M for oracle, 271M for the rule-based developer, 342M for Opus 5.5), study and running costs, and harm or safety benefit yield net value (US\$209M, 195M, 160M).}
\end{figure}

\subsection{The composition of the release}

\begin{table}[!htbp]
\caption{The four clinical programmes, one per row, differ in endpoint, price of harm, site network, and budgets. Prices of harm are per percentage point of excess risk per year; all programmes generate US\$250M per year per typical benefit unit.}
\centering\scriptsize\setlength{\tabcolsep}{2pt}
\begin{tabular}{>{\raggedright\arraybackslash}p{0.140\linewidth}>{\raggedright\arraybackslash}p{0.208\linewidth}>{\raggedright\arraybackslash}p{0.185\linewidth}>{\raggedright\arraybackslash}p{0.109\linewidth}>{\raggedright\arraybackslash}p{0.064\linewidth}>{\raggedright\arraybackslash}p{0.202\linewidth}}
\toprule
\textbf{Clinical programme} & \textbf{Endpoint} & \textbf{Typical benefit} & \textbf{Price of harm} & \textbf{Sites} & \textbf{Budgets (constrained, lean, generous)} \\
\midrule
Atopic dermatitis & Eczema Area and Severity Index & 10 points & US\$20.8M & 115 & US\$32M, 53M, 120M \\
COPD & Annual moderate or severe exacerbations & 0.25 exacerbations a year & US\$15.6M & 338 & US\$58M, 138M, 314M \\
Migraine & Monthly migraine days & 2 days per 28 & US\$12.5M & 88 & US\$34M, 48M, 109M \\
Rheumatoid arthritis & Disease Activity Score in 28 joints & 1.0 point & US\$20.8M & 139 & US\$39M, 61M, 139M \\
\bottomrule
\end{tabular}
\end{table}

\begin{table}[!htbp]
\caption{Information profiles determine how informative a programme’s evidence is, and identification settings determine how identifiable the treatment effect is from the starting cohort. Candidates are sampled as effective, partial, null, or harmful based on published phase 2 success rates, and safety ranges from protective to double the maximum acceptable excess risk.}
\centering\scriptsize\setlength{\tabcolsep}{2pt}
\begin{tabular}{>{\raggedright\arraybackslash}p{0.232\linewidth}>{\raggedright\arraybackslash}p{0.724\linewidth}}
\toprule
\textbf{Setting} & \textbf{What it changes} \\
\midrule
Standard information & Cohort 2,400 or 3,600 (3 or 5 candidates); standard dropout; 98\% of endpoint visits; measurement error 0.10 \\
High information & Cohort 3,200 or 4,800; half the dropout; all endpoint visits; no measurement error \\
Limited measurement & Cohort 1,600 or 2,400; 1.5 times the dropout; 97\% of visits; measurement error 0.15 \\
Higher attrition & Same cohort (2,400 or 3,600) and measurement error (0.10) as standard information; 1.25 times the dropout; 97\% of visits \\
Measured confounding with overlap & The effect can be recovered from the starting cohort with adjustment \\
Limited overlap & Recorded factors strongly drive who receives which treatment (channelling), so few participants are comparable and weighting loses effective sample size \\
Residual confounding & An unrecorded factor affects both treatment choice and outcomes, so the effect cannot be fully recovered from the cohort. The agent cannot establish this from the cohort alone; randomised studies can test it \\
Near tie & The benefits of the top two candidates differ by at most 0.02 of a typical effect \\
\bottomrule
\end{tabular}
\end{table}

With 4 clinical programmes, 4 information profiles, 4 identification settings, and 2 portfolio sizes, there are 128 strata, and with 3 independent worlds per stratum, there are 384 programmes. Each programme has one evidence analysis task, where the agent must report the standardised observed associations and intervals from the starting observational cohort, and 3 development tasks at different budget levels (constrained, lean, generous). Each task has both a raw and an analysis-ready view, for a total of 3,072 tasks. We reserve 64 programmes (512 tasks).

\begin{figure}[!htbp]\centering
\includegraphics[width=1.0\linewidth,height=0.88\textheight,keepaspectratio]{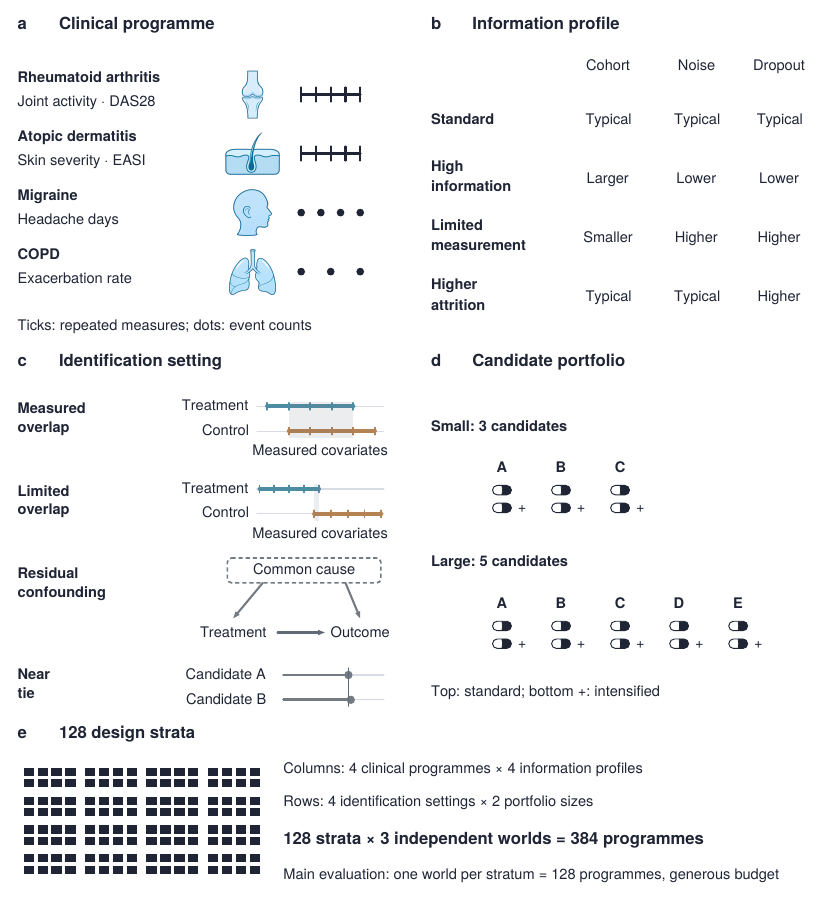}
\caption{Four crossed factors create 128 strata, with 3 independent worlds per stratum for 384 programmes. (a--d) Clinical programme, information profile, identification setting and candidate portfolio; in (c), bars span each arm's measured covariates and shading marks their overlap. (e) Each cell is one stratum; the main evaluation uses one world per stratum (128 programmes) on raw records with the generous budget.}
\end{figure}

\FloatBarrier
\section{Scoring}
\setcounter{table}{0}\setcounter{figure}{0}\renewcommand{\thetable}{\Alph{section}.\arabic{table}}\renewcommand{\thefigure}{\Alph{section}.\arabic{figure}}

\subsection{Error on replication trials}

We run each agent's saved code with no modifications on 30 replication trials and define its error on a task as the average absolute difference between its estimated effect and the true effect, divided by the standard error.

\[E_k = \frac{1}{30} \sum_{r=1}^{30} \frac{|\hat{\theta}_{kr} - \theta_k|}{s_k},\] where \(\hat{\theta}_{kr}\) is the estimate on replication trial \(r\) of task \(k\), \(\theta_k\) is the true effect and \(s_k\) is the task's standard error.

\begin{figure}[!htbp]\centering
\includegraphics[width=1.0\linewidth,height=0.9\textheight,keepaspectratio]{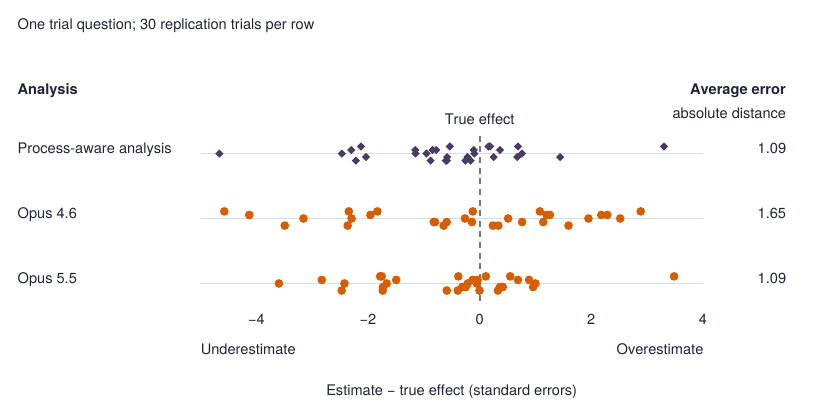}
\caption{A single trial is insufficient to evaluate an analysis, so we re-run the same saved code on replication trials. Each row shows one analysis on 30 new simulated trials per task. Each point is one trial's estimate minus the true effect, in standard errors, with overestimates to the right of the dashed line. The right column gives the average absolute distance; lower is better.}
\end{figure}

The standard error is directly measured by the variability of the estimated effect across independent simulated trials of the same size with complete data. The standard error is therefore independent of any agent's analysis and serves as a common scale across different outcome measures. If an agent does not produce a valid effect estimate or if its code fails on replication trials, we assign it the maximum between the worst model's error on that task and the error from guessing no effect, so failing to produce an answer is never better than the worst submitted answer.

\subsection{Separating the model from the difficulty of the task}

A simple averaging approach would partially measure which tasks each model has scores on. To address this, we fit \(e_{mt} = \alpha_m + \beta_t + \varepsilon_{mt}\) by least squares, where \(e_{mt}\) is the error for model \(m\) on task \(t\) and the \(\beta_t\) terms average to 0. Here \(\beta_t\) represents the relative difficulty of the task for all models, and \(\alpha_m\), the model effect, captures the model's error on an average task. We report the model effect as the model's score.

For example, if model X has errors of 0.8 and 1.2 on an easy and hard task respectively, and model Y has a score only on the easy task with an error of 0.9, a simple average would favour model Y (0.9 vs 1.0) despite Y being worse on the only task they share. Our approach yields task effects of −0.2 and 0.2, and model effects of 1.0 and 1.1, correctly ordering the models. We bootstrap confidence intervals and rankings by resampling base worlds and refitting.

\subsection{Interpreting error scores}

The process-aware analysis achieves an error score of 0.90 (95\% CI 0.87 to 0.93), which is higher than the 0.80 of an unbiased analysis because we measure the standard error on trials with complete data, whereas the process-aware analysis must operate with imperfect records. When restricted to trials without record issues, its score improves to 0.81. To outperform the process-aware analysis, an agent would need to use a valid estimator with lower variance or incorporate shrinkage towards a well-supported prior, trading off some bias for reduced variance.

Guessing no effect represents the most extreme form of shrinkage. With an error score of 1.16 (the average magnitude of the true effect), this approach only outperforms other methods when true effects are small. For effects with magnitude less than 0.5 standard errors, guessing no effect achieves a score of 0.26, compared to 0.91 for the top five models (Sonnet 5.5, Opus 5.5, Fable 5.1, Opus 5, GPT-6 Astra). For effects of 1.5 standard errors or more, guessing no effect scores 2.34, versus 1.00 for the top five models and 1.20 for the open-weight models (gpt-oss-120b, DeepSeek V4 Pro, DeepSeek V4.1 Flash, Mistral Large 4, GLM-5.3, Kimi K3, Qwen3.8 Max). The top five models stay near one standard error regardless of the true effect size, while guessing no effect fails most on the effects that trials are designed to detect.

\subsection{Five fixed analyses}

We evaluate five fixed analysis approaches on clean analysis tables: the process-aware analysis (0.90), the plan's primary analysis (0.97) which follows the primary analysis in the plan, always corrected (0.95) which always applies the analysis plan's correction, test-then-correct (0.95) which applies the correction only when the analysis plan's diagnostic detects an issue, and never corrected (1.02) which uses routine records only.

The two possible mistakes have different costs. On trials with unequal recording errors, the never corrected analysis has an average error of 1.75 compared to 1.08 for the process-aware analysis, a penalty of 0.67. On trials with equal recording errors, the always corrected analysis has an average error of 1.17 compared to 0.98 for the process-aware analysis, a penalty of 0.19, so when it is unclear whether recording errors differ, correcting is the safer choice.

\begin{table}[!htbp]
\caption{An unnecessary correction is less costly than a missed correction. A cell gives one fixed analysis’s average error on that kind of trial and, in parentheses, the fraction of those tasks that are serious misses; lower is better.}
\centering\scriptsize\setlength{\tabcolsep}{2pt}
\begin{tabular}{>{\raggedright\arraybackslash}p{0.293\linewidth}>{\raggedright\arraybackslash}p{0.178\linewidth}>{\raggedright\arraybackslash}p{0.178\linewidth}>{\raggedright\arraybackslash}p{0.168\linewidth}>{\raggedright\arraybackslash}p{0.103\linewidth}}
\toprule
\textbf{Analysis} & \textbf{No record problem} & \textbf{Unequal recording errors} & \textbf{Equal recording errors} & \textbf{All} \\
\midrule
Process-aware analysis & 0.81 (0\%) & 1.08 (0\%) & 0.98 (0\%) & 0.90 (0\%) \\
Test-then-correct & 0.81 (0\%) & 1.32 (19\%) & 1.01 (0\%) & 0.95 (4\%) \\
Never corrected & 0.81 (0\%) & 1.75 (62\%) & 0.94 (0\%) & 1.02 (12\%) \\
Always corrected & 0.81 (0\%) & 1.08 (0\%) & 1.17 (0\%) & 0.95 (0\%) \\
Plan's primary analysis & 0.81 (0\%) & 1.42 (31\%) & 0.98 (0\%) & 0.97 (6\%) \\
\bottomrule
\end{tabular}
\end{table}

\subsection{Development value}

To illustrate how the components in Section 4 combine, consider a migraine drug with a benefit of 3 fewer migraine days per 28 days and an excess risk of 2 percentage points. Advanced on day 1,500 after US\$50 million in prior studies, this drug generates US\$1,711 million in discounted profit, incurs US\$114 million in cost of harm, and US\$22.5 million in operating costs, yielding a value of US\$1,524 million.

\begin{table}[!htbp]
\caption{A drug’s value is far higher when advanced earlier: for migraine, a drug with 3 fewer migraine days per 28 days is worth US\$2,870M on day 0 versus US\$1,711M on day 1,500. Each cell shows the discounted profit from launch to exclusivity end, before the cost of harm and other expenses.}
\centering\scriptsize\setlength{\tabcolsep}{2pt}
\begin{tabular}{>{\raggedright\arraybackslash}p{0.212\linewidth}>{\raggedright\arraybackslash}p{0.116\linewidth}>{\raggedright\arraybackslash}p{0.116\linewidth}>{\raggedright\arraybackslash}p{0.182\linewidth}>{\raggedright\arraybackslash}p{0.141\linewidth}>{\raggedright\arraybackslash}p{0.141\linewidth}}
\toprule
\textbf{True effect} & \textbf{Benefit units} & \textbf{Yearly profit} & \textbf{Worth if advanced on day 0} & \textbf{On day 1,500} & \textbf{On day 2,190} \\
\midrule
1 fewer migraine day per 28 days & 0.5 & US\$125M & US\$957M & US\$570M & US\$437M \\
2 fewer migraine days per 28 days & 1 & US\$250M & US\$1,913M & US\$1,141M & US\$875M \\
3 fewer migraine days per 28 days & 1.5 & US\$375M & US\$2,870M & US\$1,711M & US\$1,312M \\
\bottomrule
\end{tabular}
\end{table}

\[V = \mathbf{1}_{\mathrm{approved}}\,(B - S) - \sum_j C_j - 15{,}000\,T,\] where \(B\) and \(S\) are the discounted clinical value and cost of harm over the launch window at the true effects, with the brief's dollar weights, \(C_j\) are study commitments and \(T\) is the terminal day. Stopping on day 0 earns zero, and an unfinished run uses its final day as \(T\).

A drug candidate only realises this value upon approval. The approval review aggregates all confirmation studies for the advanced candidate and regimen with elapsed follow-up time, using complete outcome data (including outcomes missing from records) for all randomised participants in both treatment and control arms. Approval requires statistically significant evidence of benefit at the two-sided 5\% level with at least 300 completed participants in the treatment arm. Value is then calculated based on the true effect across all 64,000 participants, regardless of which participants enrolled. This means a drug can be approved by chance due to a lucky sample, but cannot have inflated value.

\subsection{The oracle}

For each candidate and regimen, the oracle searches across 6 site fraction levels, 8 confirmation study sizes, and 2 follow-up times, selecting the plan with highest expected net value: the average value across possible trial outcomes minus costs. If no plan has positive expected net value, it stops on day 0. Otherwise, it executes the smallest permitted early safety and exploration studies, then conducts the confirmation trial, advancing without responding to any results.

\subsection{The rule-based developer}

The rule-based developer (RBD) receives the same information as an agent: the starting cohort, study results, study costs, the brief, and site predictions (public forecasts of site enrolment rates and study timelines, not drug efficacy).

The RBD also uses a prespecified statistical model with a prior from Cochrane effect sizes, fitted in advance on 192 separate simulated drug development programmes (48 per disease area) and then frozen. The model captures the distribution of candidate drug effects on benefit and safety, how observational and early studies may over- or underestimate these effects, and how effects generalise to confirmation trials. A study is affordable if its forecast full cost at the public phase prices in Table A.4, including setup, sites, screening, enrolment, follow-up and visits, fits within the budget left after money already spent and costs owed for enrolled participants.

The RBD only screens standard regimens and makes seven decisions in sequence.

\begin{figure}[!htbp]\centering
\includegraphics[width=1.0\linewidth,height=0.9\textheight,keepaspectratio]{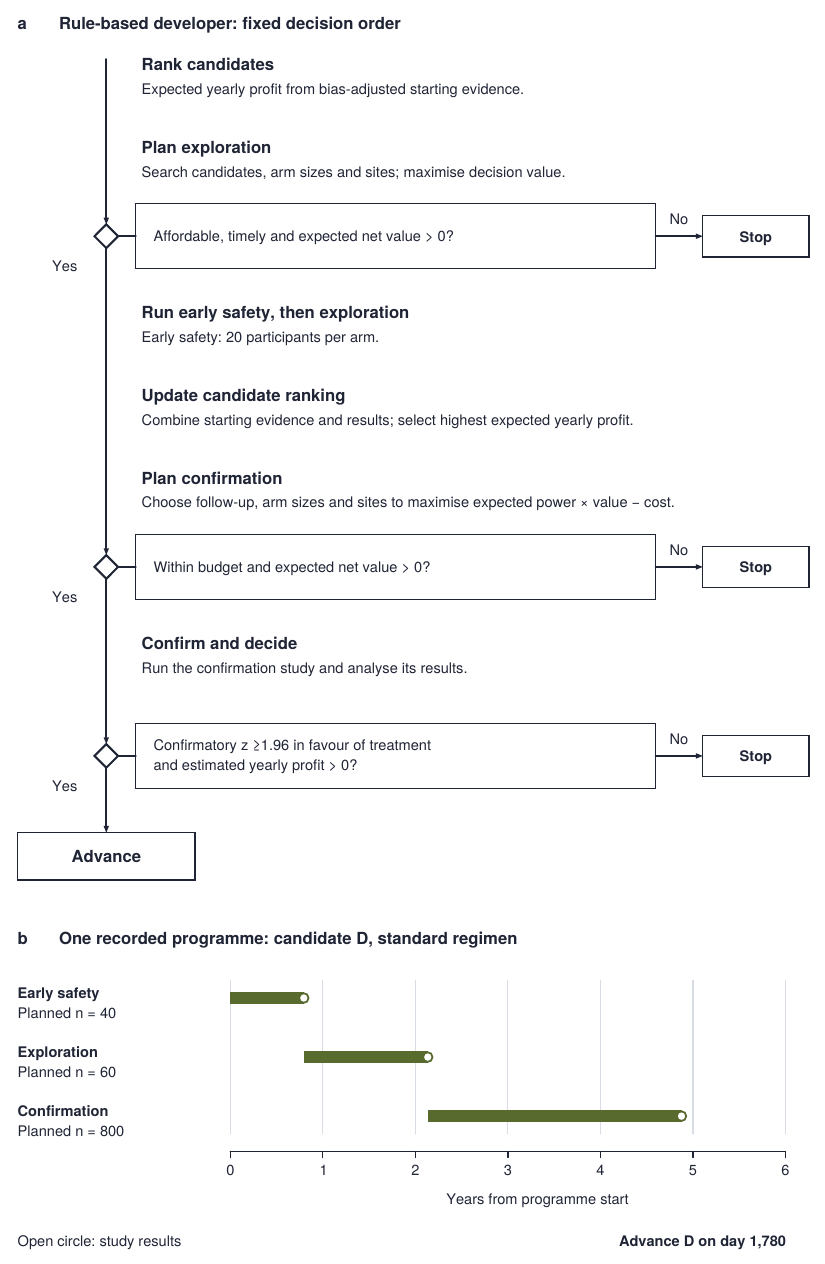}
\caption{The rule-based developer proceeds decision-by-decision in order and stops if the next step is unaffordable, untimely, or has negative expected value. (a) The order of decisions with each check being a diamond shape; if any of these checks fail, the programme is stopped. (b) One example programme with the planned number of participants enrolled per study and the advance day.}
\end{figure}

\begin{table}[!htbp]
\caption{The RBD’s seven decisions, in the order they are made.}
\centering\scriptsize\setlength{\tabcolsep}{2pt}
\begin{tabular}{>{\raggedright\arraybackslash}p{0.167\linewidth}>{\raggedright\arraybackslash}p{0.789\linewidth}}
\toprule
\textbf{Rule} & \textbf{Action} \\
\midrule
1. Budget & REQUIRE total cost of the plan within budget. Compare feasible plans by expected return after costs and delay. \\
2. Rank & FOR each candidate: estimate its association from the starting cohort, adjust it with the fitted model, compute expected yearly profit after harm. SORT candidates by this profit. \\
3. Plan exploration & FOR each plan (leading candidates, participants per arm, sites): average over the results it could return, the update each would cause and the later stop-or-confirm decision; subtract study costs. IF no affordable, timely plan has expected net value above 0: STOP on day 0. ELSE choose the highest. \\
4. Run studies & RUN early safety (20 participants per arm; allowed minimum 10), THEN the chosen exploration study. Standard regimens only; one results release per study; no adaptations. \\
5. Update & COMBINE the starting association, the fitted model and the exploration result. KEEP the candidate with the highest expected yearly profit after harm. \\
6. Confirm & FOR each design (follow-up, size, sites) within the remaining budget: expected net value = probability of passing review $\times$ value if approved − study and time costs. IF the best is not above 0: STOP. ELSE run it. \\
7. Decide & IF z $\geq$ 1.96 in favour of treatment (two-sided 5\%) AND estimated yearly profit above 0: ADVANCE. ELSE STOP. \\
\bottomrule
\end{tabular}
\end{table}

For decisions 3 and 6, which involve planning studies before results are available, the RBD must consider all possible outcomes and subsequent decisions. For example, if a plan has a 0.6 probability of yielding US\$200 million in approval value with an additional US\$30 million in costs, its expected net value is 0.6 × 200 - 30 = US\$90 million.

\FloatBarrier
\section{Validity}
\setcounter{table}{0}\setcounter{figure}{0}\renewcommand{\thetable}{\Alph{section}.\arabic{table}}\renewcommand{\thefigure}{\Alph{section}.\arabic{figure}}

Differences between models should stem from the models themselves rather than the benchmark design.

\begin{figure}[!htbp]\centering
\includegraphics[width=1.0\linewidth,height=0.76\textheight,keepaspectratio]{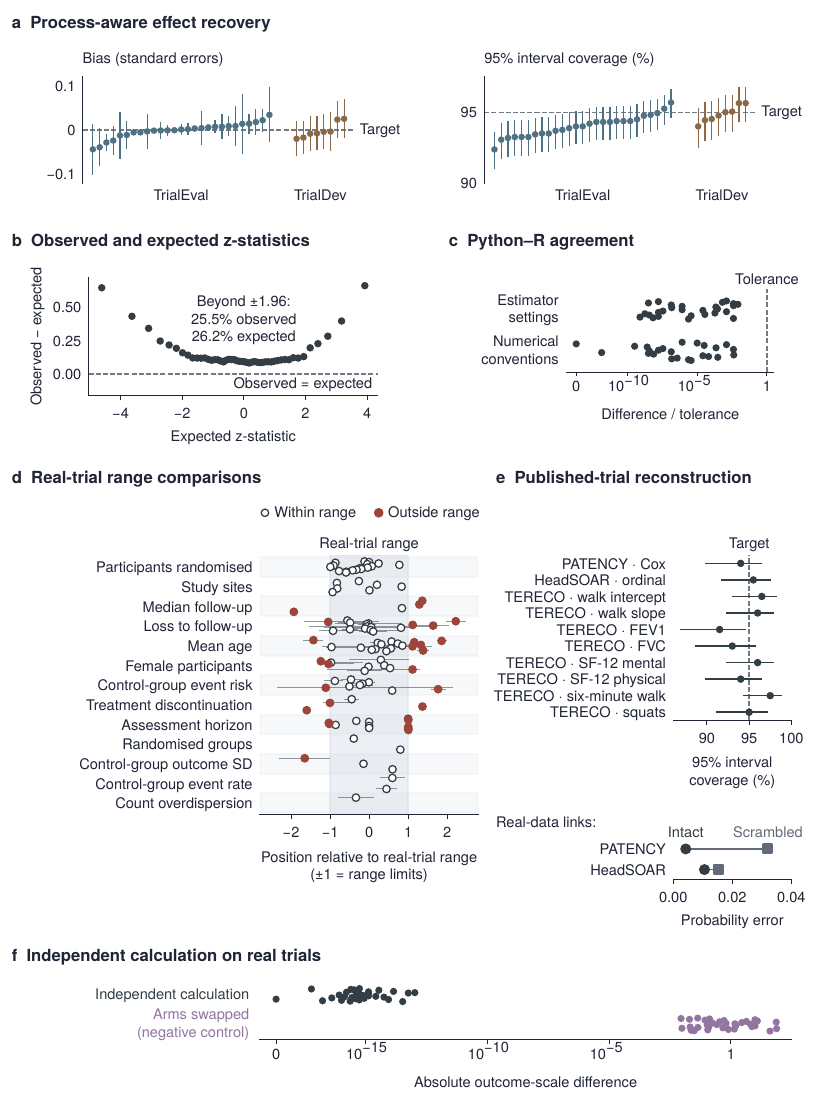}
\caption{Process-aware analysis recovers the specified effect and the analysis code is reproducible independently. (a) Bias and 95\% interval coverage for each configuration: bias remains close to zero and coverage close to 95\%. (b) Observed minus expected z-statistic across simulated trials: about 0.1 higher than expected in the middle and up to 0.65 higher in the tails; over $\pm$1.96, 25.5\% observed vs 26.2\% expected. (c) Maximum difference between the Python and R implementations relative to the tolerance: all settings agree within 1\% of tolerance. (d) Properties of simulated questions compared to the range of real trials: 75 of 103 comparisons are within the range. (e) True effect coverage for three re-simulated published trials is close to 95\% except for TERECO FEV1 (forced expiratory volume in one second), and scrambling real data links increases probability error. (f) Our analysis of 33 real randomised trials agrees with an independent calculation to floating point precision, whereas swapping treatment arms leads to large differences.}
\end{figure}

\subsection{Recovery of the specified effect}

Since we control the data generation process and set the true effect for each world, we can verify whether process-aware analysis can accurately recover this true effect. We run many simulated trials for each configuration and compute the average estimates. Bias is measured as the difference between the average estimate and the true effect, normalised by the standard error. On TrialEval, where true effects are randomly sampled from the Cochrane-informed distribution within each question's range, we observe that bias remains below 0.05 standard errors across all configurations. Coverage, which measures how often the 95\% confidence intervals contain the true effect, ranges between 92\% and 96\% when true effects are fixed at their release values. On TrialDev, we observe the same range of coverage on confirmation trials across all endpoints when using TrialDev's own true effect distribution, which is informed by published phase 2 success rates.

These checks would be meaningless if recording errors were negligible. When recording errors are unequal, we find that the standard analysis deviates by at least 1.0 standard errors for all true effects tested.

\subsection{Realism against real trials}

For each trial characteristic, we exclude the registry trials and published examples used to calibrate it. We then compare the simulated trials against the remaining real-world trials.

\begin{figure}[!htbp]\centering
\includegraphics[width=1.0\linewidth,height=0.9\textheight,keepaspectratio]{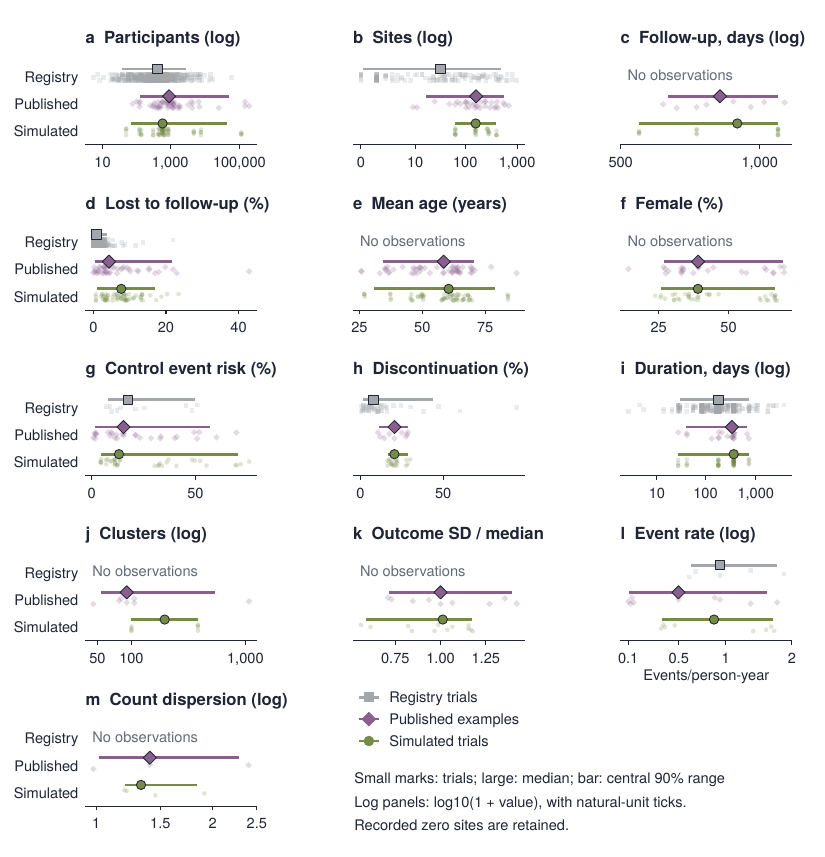}
\caption{Simulated trials are within or close to the range of real trials. Each panel shows one characteristic: registry trials (squares), published examples (diamonds), and simulated trials (circles). Small marks are individual observations; the large mark is the median and the bar is the central 90\% range. Outcome SD is relative to the published-example median in each clinical setting.}
\end{figure}

True effect sizes also align with real-world data. A z-statistic (estimate divided by standard error) with absolute value greater than 1.96 indicates that the 95\% confidence interval excludes zero. For process-aware analysis, 25.5\% of z-statistics exceed this threshold, slightly below the 26.2\% expected (Figure C.1b) from the Cochrane-informed distribution. The signal strength agents need to detect is therefore comparable to that in real trials. The clinical questions are also similar in size, outcome time points, and number of sites to held-out registry trials.

On TrialDev, cohorts match their target profiles and identification settings change the strength of the starting evidence as intended, so agents must assess how informative their starting evidence is before deciding what to purchase.

\begin{figure}[!htbp]\centering
\includegraphics[width=1.0\linewidth,height=0.9\textheight,keepaspectratio]{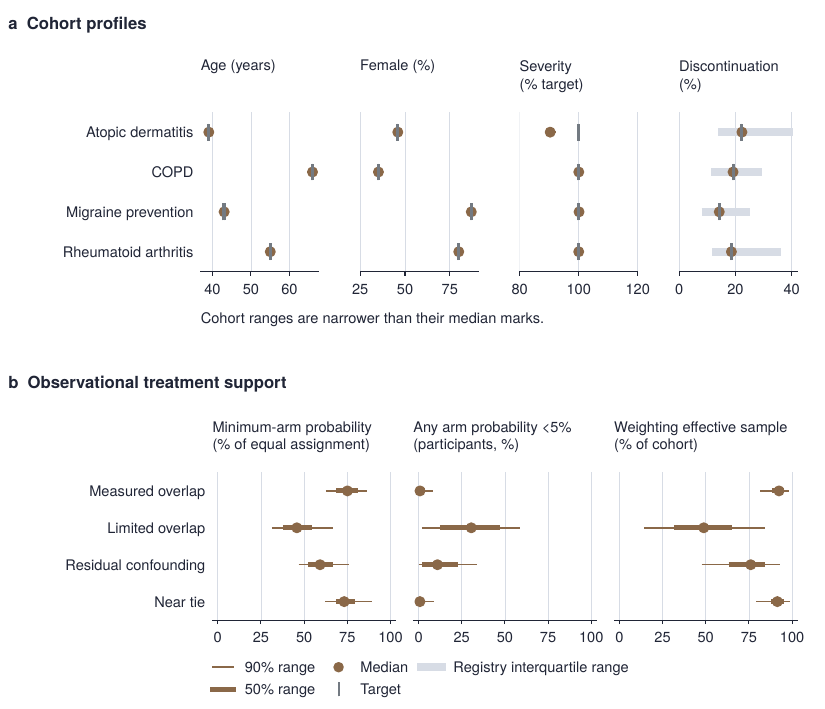}
\caption{TrialDev cohorts align with their targets, and starting evidence is weakest with limited overlap. (a) Cohort profiles by clinical programme: medians, central 50\% and 90\% ranges, and profile targets. Registry bars show interquartile ranges for discontinuation. (b) Treatment support by identification setting: how close the smallest treatment probability is to equal assignment, the share of participants with any treatment option below 5\% probability, and the effective sample size left after weighting.}
\end{figure}

\subsection{Reproduction by independent calculation}

We verify that all scored estimators match an independent R implementation on generated data: the point estimate, standard error, and both confidence interval bounds all fall within a small numerical tolerance. For real randomised trials, we confirm that our benchmark analysis code matches an independent calculation up to rounding errors and correctly flags all negative control trials where we swap treatment arms.

We additionally re-simulated 3 published trials (PATENCY, HeadSOAR, TERECO) \cite{tian2025patency, yuan2026headsoarbmj, yuan2026headsoar, li2022tereco}: we specified each trial's design and population in the simulator, drew new trials from it with a known true effect, and measured how often the 95\% confidence interval contains the true effect used for simulation: PATENCY (0.94), HeadSOAR (0.955), and TERECO (0.915 to 0.975 across outcomes). On TrialDev, all decisions and values can be reproduced by an independent verifier.

\subsection{Reliability of model rankings}

To assess the reliability of model rankings, we randomly split the task set while keeping different versions of the same base world together. We then score all models on each half, calculate the correlation between scores, and repeat this process. The Spearman--Brown split-half reliability is 0.895 for both task suites (where 1 represents perfect correlation), so model rankings depend little on which tasks are sampled.

\begin{figure}[!htbp]\centering
\includegraphics[width=1.0\linewidth,height=0.9\textheight,keepaspectratio]{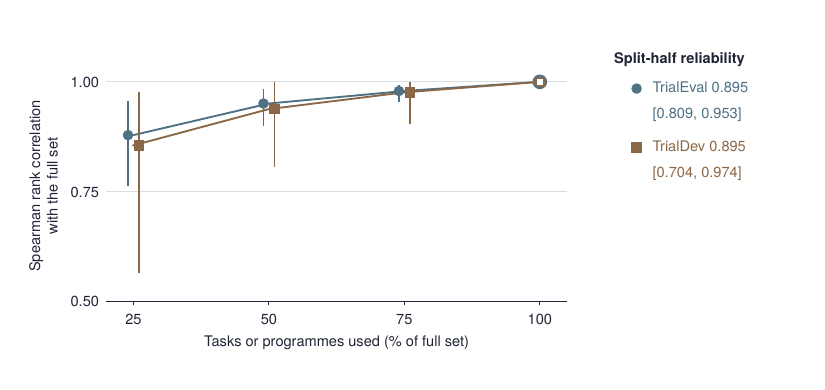}
\caption{Model rankings are similar using two random halves of tasks and converge rapidly. Spearman rank correlation with the full task set for random stratified subsets of 25\%, 50\% and 75\% of tasks. Whiskers indicate the range of correlations for repeated subsets. Spearman–Brown split-half reliability for the full task set and the range of split-half reliabilities across random splits are annotated next to the curves.}
\end{figure}

\FloatBarrier
\section{Agent behaviour in analysis}
\setcounter{table}{0}\setcounter{figure}{0}\renewcommand{\thetable}{\Alph{section}.\arabic{table}}\renewcommand{\thefigure}{\Alph{section}.\arabic{figure}}

The top models are good at cleaning raw records. The key difference from the process-aware analysis is when to apply data corrections.

\begin{figure}[!htbp]\centering
\includegraphics[width=0.62\linewidth,height=0.9\textheight,keepaspectratio]{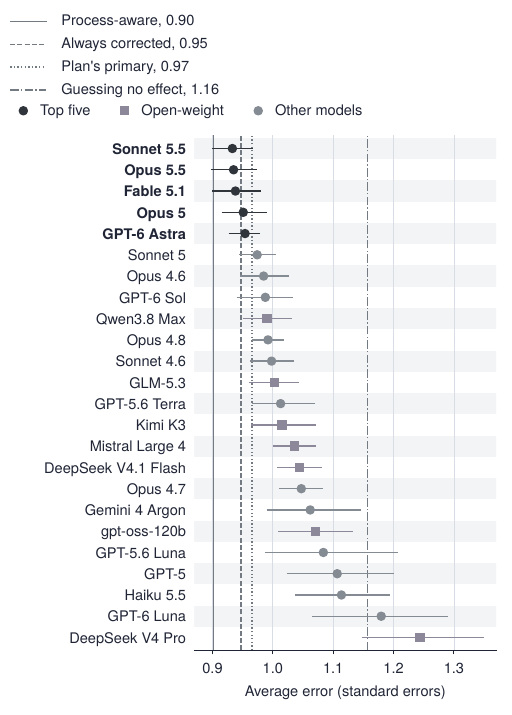}
\caption{Model errors range from close to the process-aware analysis to higher than the error of guessing no effect. Average error for each model on the main tasks with the 95\% interval from bootstrap resampling. Lower error is better. Vertical lines indicate the error of the process-aware analysis, always corrected, the plan's primary analysis, and guessing no effect.}
\end{figure}

\subsection{Preparing raw records for analysis}

Before estimating the treatment effect, agents must clean the raw source tables into a single analysis table. This involves removing extra records, deduplicating records by retaining only the latest version, converting non-day time units, and harmonising identifiers. We search through the final scripts of each model for code snippets that implement these cleaning steps on tasks that require them. For the top five models, the fractions of scripts that contain code for these four steps are 87\%, 94\%, 67\%, and 100\%, respectively. For the open-weight models, the fractions are 70\%, 86\%, 59\%, and 92\%.

To test whether these steps are correct, we turn to accuracy. For tasks that do not have record issues where a valid answer only requires data cleaning and a standard analysis, the top five models have an error of 0.81, which is the same as the process-aware analysis on clean tables (difference 0.00, 95\% CI −0.02 to 0.02). The open-weight models have an error of 0.86, partly because some of their runs return no usable answer.

\subsection{Correcting by default}

A script's code only shows whether it attempts to perform data correction, so we further examine the behaviour of the script's estimates. For each model and task, we compute the script's estimate \(a\) across replication trials, the never corrected estimate \(u\), and the always corrected estimate \(c\). We then compute the correction index as \(\sum(a-u)(c-u) / \sum(c-u)^2\), clipped to the range −0.5 to 1.5. A correction index of 0 indicates that the script's estimates behave like the never corrected estimates, while an index of 1 indicates that they behave like the always corrected estimates. For example, if \(u = 0\), \(c = 2\), and the script's estimates on all replication trials are 1.8, then the correction index is 0.9. We say that a script behaves as corrected if the correction index is above 0.5, i.e., its estimates are closer in squared distance to the always corrected estimates than to the never corrected estimates.

The top five models behave as corrected on 99\% of tasks with unequal recording errors, where correction is beneficial, and on 83\% of tasks with equal recording errors, where correction is detrimental. In contrast, the process-aware analysis behaves as corrected on 100\% and 5\% (1 out of 20 tasks) of these tasks, respectively: it uses the arm-specific correction only when recording errors are unequal, and otherwise follows the plan, which can itself draw on the audit. The open-weight models behave as corrected on 49\% and 36\% of these tasks.

To summarise how well a model can tell these two types of tasks apart, we use d′: how well its decision to correct separates the tasks that need correction from those that do not, regardless of how often it corrects overall. Formally, \(d' = \Phi^{-1}(H) - \Phi^{-1}(F)\), where \(\Phi^{-1}\) is the inverse of the cumulative distribution function of the standard normal distribution, \(H\) is the hit rate (fraction of tasks with unequal recording errors where the model applies correction), and \(F\) is the false positive rate (fraction of tasks with equal recording errors where the model applies correction). A d′ of 0 corresponds to random chance. The d′ for the top five models is 1.3, for all models is 0.36, and for the open-weight models is 0.33.

The default correction behaviour is a sound strategy given the cost difference in Appendix B. The top five models have similar error to the process-aware analysis on tasks with unequal recording errors (1.09 vs 1.08), and their error on tasks with equal recording errors is close to the always corrected estimates (1.15 vs 1.17). Overall, the top five models have similar error to the test-then-correct estimates (difference in error −0.01).

\begin{figure}[!htbp]\centering
\includegraphics[width=1.0\linewidth,height=0.9\textheight,keepaspectratio]{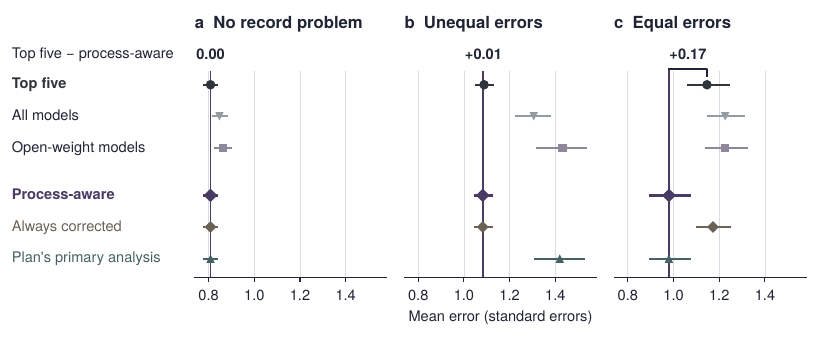}
\caption{The top five models match the process-aware analysis when there is no record problem or recording errors are unequal, but trail it by 0.17 standard errors when errors are equal, close to always correcting. Each row is the mean error of a model group or fixed analysis with its 95\% interval; the purple line marks the process-aware analysis, and the numbers above each column give the top five's difference from it.}
\end{figure}

However, the correction decisions of the models are inconsistent. The models should treat all worlds within a scenario consistently, yet the top five models exhibit inconsistent behaviour across worlds within a scenario on 36\% of model-scenario pairs with equal recording errors and 5\% with unequal recording errors. Since the models check for data discrepancies in 83--100\% of the runs, they have access to the relevant information but lack a reliable way to interpret it.

\subsection{Serious misses}

We classify the 196 serious misses across all models into different failure modes based on the task type, whether the model's script attempts a correction, and whether it returns a usable answer.

\begin{table}[!htbp]
\caption{Most of the 196 serious misses are related to the decision to correct. Each row represents a failure category, and each serious miss belongs to exactly one category. Models and Tasks show counts of those with at least one serious miss in that category.}
\centering\scriptsize\setlength{\tabcolsep}{2pt}
\begin{tabular}{>{\raggedright\arraybackslash}p{0.424\linewidth}>{\raggedright\arraybackslash}p{0.160\linewidth}>{\raggedright\arraybackslash}p{0.119\linewidth}>{\raggedright\arraybackslash}p{0.119\linewidth}>{\raggedright\arraybackslash}p{0.097\linewidth}}
\toprule
\textbf{Failure} & \textbf{Serious misses} & \textbf{Share, \%} & \textbf{Models} & \textbf{Tasks} \\
\midrule
Correction with equal errors & 66 & 34 & 20 & 9 \\
No usable answer & 48 & 24 & 11 & 31 \\
Correction attempted with unequal errors, still far from the truth & 41 & 21 & 18 & 9 \\
No correction with unequal errors & 26 & 13 & 9 & 7 \\
Valid script that fails on replication trials & 9 & 5 & 7 & 8 \\
No record problem, far from the truth & 6 & 3 & 6 & 6 \\
\bottomrule
\end{tabular}
\end{table}

\begin{figure}[!htbp]\centering
\includegraphics[width=1.0\linewidth,height=0.9\textheight,keepaspectratio]{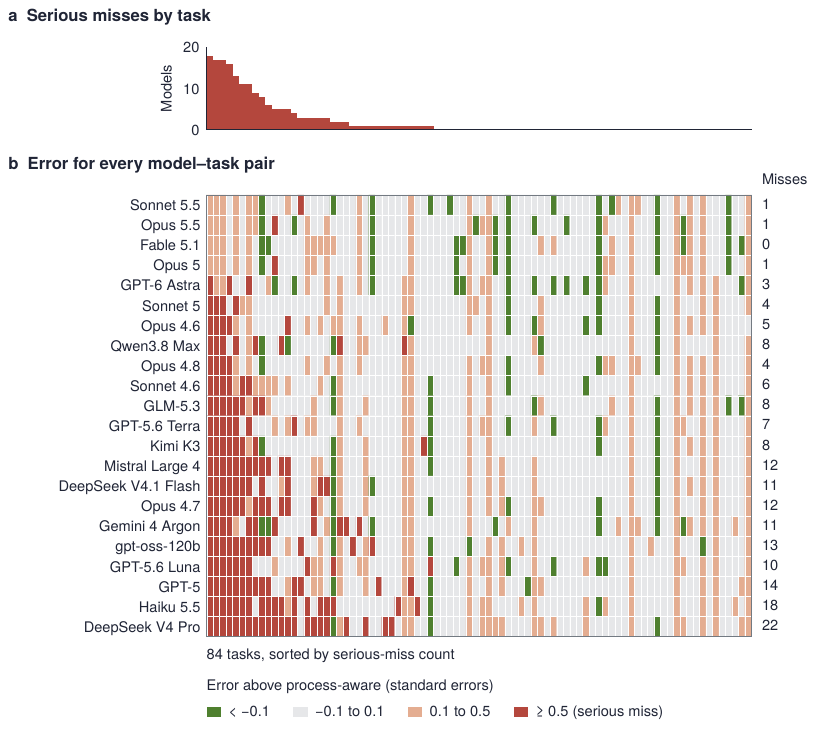}
\caption{Serious misses concentrate on a minority of tasks. Across the 22 models evaluated on all 84 tasks, the 179 serious misses fall on 35 tasks and the other 49 tasks have none; each of the top five models has at most 3, while the weakest has 22. (a) Number of models with a serious miss on each task. (b) Each cell is one model on one task, coloured by its error above the process-aware analysis in standard errors; a serious miss is an error at least 0.5 above. Models are ordered by overall TrialEval error and tasks by serious-miss count.}
\end{figure}

\begin{figure}[!tp]\centering
\includegraphics[width=1.0\linewidth,height=0.9\textheight,keepaspectratio]{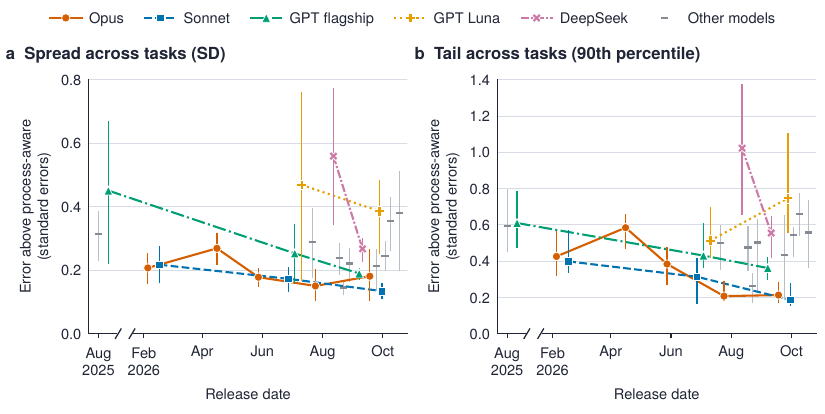}
\caption{The tail of worst-performing tasks shortens within the Opus, Sonnet and flagship GPT lines. Spread (a, standard deviation) and tail (b, 90th percentile) of each release's error above the process-aware analysis across tasks, with 95\% intervals; lines join successive releases of a model line and grey marks are other models. From first to latest release, the 90th percentile falls from 0.43 to 0.21 for Opus, 0.40 to 0.19 for Sonnet and 0.61 to 0.36 for flagship GPT.}
\end{figure}

\subsection{Plans and context levels}

Analysis plans and improved data input do not significantly improve model performance. When given a fixed analysis plan, the models diverge from the plan on most tasks: 54\% for tasks without record issues, 66\% for tasks with equal recording errors, and 71\% for tasks with unequal recording errors where following the plan's primary analysis would harm the model's accuracy, so a locked plan does not make agents follow it, even where its primary analysis is as accurate as the process-aware analysis. Across 3 models, we do not detect a clear average penalty in error for using raw records instead of analysis-ready tables (−0.03, 95\% CI −0.11 to 0.04). Providing models with a list of known data defects as context does not significantly change their error (+0.06, 95\% CI −0.03 to 0.15).

\begin{table}[!htbp]
\caption{Locking the plan or using cleaner tables results in minimal change in error for the three models evaluated on these context levels. Each cell shows a model’s raw mean error in standard errors (lower is better); dash indicates level not evaluated.}
\centering\scriptsize\setlength{\tabcolsep}{2pt}
\begin{tabular}{>{\raggedright\arraybackslash}p{0.197\linewidth}>{\raggedright\arraybackslash}p{0.142\linewidth}>{\raggedright\arraybackslash}p{0.142\linewidth}>{\raggedright\arraybackslash}p{0.142\linewidth}>{\raggedright\arraybackslash}p{0.142\linewidth}>{\raggedright\arraybackslash}p{0.142\linewidth}}
\toprule
\textbf{Model} & \textbf{C1} & \textbf{C2} & \textbf{C3} & \textbf{C4} & \textbf{C5} \\
\midrule
Sonnet 5.5 & 0.94 & -- & 1.01 & -- & 0.93 \\
Haiku 5.5 & 1.14 & 1.16 & 0.97 & 1.04 & 1.11 \\
GPT-6 Luna & 1.17 & 1.08 & 1.18 & 1.15 & 1.19 \\
\bottomrule
\end{tabular}
\end{table}

\subsection{Changes across model releases}

Within model lines, the main improvement is in the worst-case performance. From the earliest to the latest model version, the 90th percentile of analysis error above the process-aware analysis decreases by 0.21 for Opus, 0.21 for Sonnet, and 0.25 for the flagship GPT. The serious miss rate is 14\% for Opus 4.7 and 1\% for Opus 5.5.

Model improvement is not monotonic across releases. Opus 4.7 is worse than Opus 4.6 and GPT-6 Luna is worse than GPT-5.6 Luna, so a newer release cannot be assumed to analyse trials better than its predecessor.

\FloatBarrier
\section{Decision experiments}
\setcounter{table}{0}\setcounter{figure}{0}\renewcommand{\thetable}{\Alph{section}.\arabic{table}}\renewcommand{\thefigure}{\Alph{section}.\arabic{figure}}

To determine where agents lose value over long horizons, we replay the recorded programmes on the frozen pool, modify a single element, and then re-score. The frozen pool contains all results under all treatments, so a modified programme trajectory can be computed without additional model calls. Replaying the programmes without modification exactly replicates the review results and value of all programmes that the agents advanced. Rerunning the RBD exactly replicates the recorded value of the RBD. All confidence intervals are 95\% bootstrap intervals over worlds. Unless specified, all results are reported across all recorded agent programmes.

\begin{table}[!htbp]
\caption{On average, agents run more and larger studies than the RBD and oracle, yet capture less value. Each cell shows the mean per programme under the generous budget. Value is expressed as a percentage of oracle value. Right candidate is the percentage of oracle-advanced programmes on which the developer advanced the oracle’s candidate.}
\centering\fontsize{6.8}{7.6}\selectfont\setlength{\tabcolsep}{2pt}
\begin{tabular}{@{}lrrrrrrrrrrrrr@{}}
\toprule
\textbf{\shortstack[l]{Developer}} & \textbf{\shortstack[l]{Value,\\\%}} & \textbf{\shortstack[l]{Clinical\\value, pts}} & \textbf{\shortstack[l]{Cost of\\harm, pts}} & \textbf{\shortstack[l]{Studies}} & \textbf{\shortstack[l]{Arms}} & \textbf{\shortstack[l]{Partic.}} & \textbf{\shortstack[l]{Cost,\\US\$M}} & \textbf{\shortstack[l]{Result\\releases}} & \textbf{\shortstack[l]{Turns}} & \textbf{\shortstack[l]{Adv.,\\\%}} & \textbf{\shortstack[l]{Appr.,\\\%}} & \textbf{\shortstack[l]{Intens.,\\\%}} & \textbf{\shortstack[l]{Right\\cand., \%}} \\
\midrule
Opus 5.5 & 48 & 158.1 & 80.4 & 3.2 & 4.3 & 1406 & 79 & 3.7 & 36 & 80 & 71 & 54 & 71 \\
Kimi K3 & 34 & 102.3 & 37.1 & 3.1 & 3.8 & 1653 & 80 & 3.3 & 51 & 68 & 51 & 45 & 64 \\
Sonnet 5.5 & 29 & 90.2 & 33.8 & 2.7 & 3.6 & 1491 & 72 & 2.7 & 26 & 61 & 45 & 32 & 53 \\
GPT-6 Sol & 25 & 102.2 & 28.8 & 3.8 & 3.5 & 2388 & 121 & 6.2 & 78 & 69 & 47 & 36 & 69 \\
GPT-6 Astra & 14 & 81.0 & 20.0 & 3.6 & 5.3 & 2474 & 131 & 5.6 & 87 & 53 & 45 & 28 & 43 \\
Haiku 5.5 & 11 & 54.5 & 24.0 & 2.8 & 2.3 & 928 & 42 & 2.8 & 37 & 43 & 29 & 24 & 41 \\
GLM-5.3 & 9 & 72.1 & 35.2 & 2.8 & 4.0 & 1413 & 74 & 3.2 & 68 & 47 & 36 & 58 & 41 \\
Gemini 4 Argon & 2 & 38.6 & 15.2 & 2.8 & 3.7 & 1019 & 55 & 2.8 & 99 & 45 & 19 & 46 & 35 \\
Qwen3.8 Max & −6 & 18.6 & 9.5 & 2.2 & 3.8 & 660 & 39 & 1.8 & 47 & 22 & 9 & 50 & 20 \\
GPT-6 Luna & −10 & 7.9 & 3.6 & 1.8 & 2.4 & 656 & 32 & 1.6 & 97 & 12 & 5 & 12 & 10 \\
Pooled & 17 & 74.7 & 30.3 & 2.9 & 3.8 & 1396 & 72 & 3.3 & 60 & 51 & 37 & 41 & 45 \\
Rule-based developer & 58 & 96.7 & 18.2 & 2.4 & 1.1 & 731 & 51 & 2.4 & -- & 56 & 54 & 0 & 52 \\
Oracle & 100 & 154.7 & 33.9 & 2.4 & 1.0 & 787 & 52 & 2.4 & -- & 81 & 70 & 38 & 100 \\
\bottomrule
\end{tabular}
\end{table}

\subsection{Where value is lost}

The agents achieve 16.7\% of the oracle value, while the RBD achieves 57.5\% on the same programmes, a gap of 40.8 points, equivalent to US\$139 million per programme. To pinpoint where value is lost, we run the RBD starting at progressively later phases, using the agents' programme. This creates five versions of each programme: (1) the RBD starting from day 0, (2) the agent's early safety studies followed by the RBD, (3) the agent's studies up to and including exploration, where the RBD then selects a treatment for confirmation using the agent's estimates from exploration, (4) the agent's entire programme except the final decision, which is replaced with a fixed procedure as described in the next subsection, and (5) the agent's programme as is. For versions (1) through (3), the RBD is run with the remaining budget and time and can recruit from the unused participant pool.

We attribute value loss between consecutive versions to the phase between them. For example, if a programme is valued at US\$300 million, US\$280 million, US\$100 million, US\$60 million, and US\$50 million across the five versions, we attribute US\$20 million in value loss to early safety, US\$180 million to exploration, US\$40 million to confirmation, and US\$10 million to final decision-making. These losses sum to the US\$250 million gap between the RBD and the agent. Programmes that did not finish are treated as a separate category, with the full value gap attributed to them.

On aggregate, the values of these five programme versions as a percentage of the oracle value are 57.5\%, 45.6\%, 27.6\%, 17.2\%, and 16.7\%. This allows us to attribute value loss to exploration (18.0 points, 12.3 to 24.1), confirmation (10.4, 4.8 to 17.1), unfinished programmes (9.2, 6.9 to 11.5), early safety (2.7, −2.0 to 7.2; the first step also contains unfinished programmes), and final decision (0.5, −1.8 to 2.6). Of the value loss attributed to exploration, 16.6 points are due to clinical value, indicating that following the agent's exploration phase, the RBD cannot achieve as high a clinical value as it could after its own exploration. For Opus 5.5, there is no clear loss during this phase (1.0 point, −9.1 to 11.1), so its shortfall arises after exploration. The RBD has access to prior information from its prespecified statistical model that the agents do not have. Thus, the value loss attributed to these phases represents an upper bound on what can be recovered through better judgement alone.

\begin{figure}[!htbp]\centering
\includegraphics[width=1.0\linewidth,height=0.9\textheight,keepaspectratio]{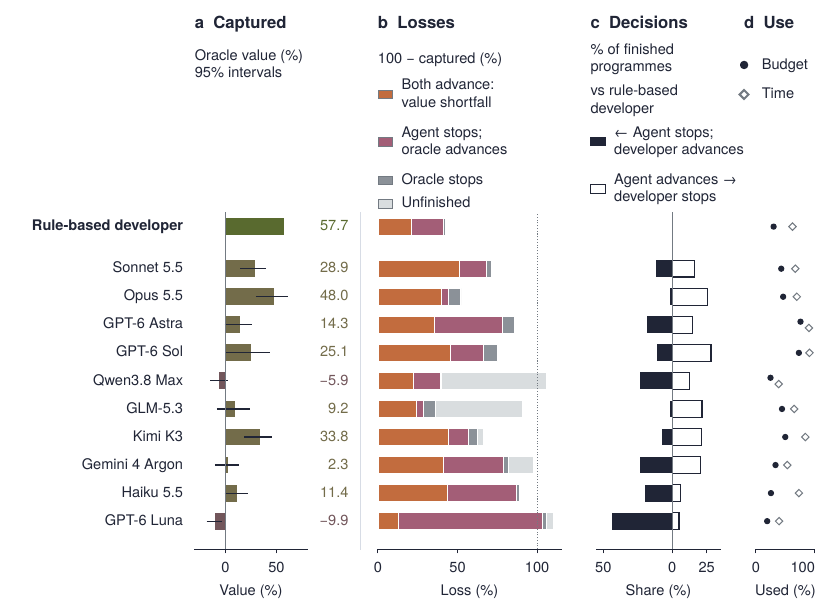}
\caption{Agents leave much of the oracle's value uncaptured, and lose it differently. (a) Signed captured value as a percentage of the oracle total, with agent 95\% intervals. (b) Losses where the agent and oracle both advance, the agent stops while the oracle advances, the oracle stops, or the programme is unfinished. (c) Among finished programmes, the fraction where the agent stops but the rule-based developer advances (solid, left), or vice versa (open, right). (d) Mean budget and calendar use. The separated first row is the rule-based developer.}
\end{figure}

We can decompose the same 40.8 points by considering how the agent and RBD end the programme. The largest source of value loss, 15.3 points, comes from programmes where the RBD achieved approval but the agent stopped. Half of all stops occurred before any confirmation study, and two-thirds of the others had fewer than 300 completed treated participants for their most significant confirmed treatment, so this treatment could not have been approved. Another 10.4 points come from programmes where the RBD achieved approval but the agent advanced a treatment that failed review.

\begin{table}[!htbp]
\caption{The largest fraction of the 40.8-point value loss comes from cases where the RBD achieved approval but the agent stopped. Each row groups the agent programmes by their final outcome for both agent and RBD. Points with 95\% confidence intervals show the RBD’s value minus the agent’s value within the group, normalised by the oracle’s total value; negative values indicate the agent performed better.}
\centering\scriptsize\setlength{\tabcolsep}{2pt}
\begin{tabular}{>{\raggedright\arraybackslash}p{0.578\linewidth}>{\raggedright\arraybackslash}p{0.378\linewidth}}
\toprule
\textbf{Ending} & \textbf{Points (95\% interval)} \\
\midrule
Both approved & 7.5 (4.2 to 11.3) \\
RBD approved; agent advanced, not approved & 10.4 (7.6 to 13.7) \\
RBD approved; agent stopped & 15.3 (11.7 to 19.4) \\
Agent approved; RBD not & −4.9 (−9.0 to −1.1) \\
Neither approved & 3.4 (2.0 to 5.2) \\
Agent unfinished & 9.2 (6.9 to 11.5) \\
\bottomrule
\end{tabular}
\end{table}

Spending is not the main source of this gap: of the 40.8 points, only 6.5 are due to study and time costs, while 22.2 are due to clinical value and 12.1 are due to cost of harm.

\subsection{Same evidence, fixed final decision}

To isolate the impact of the final decision, we retain all studies conducted by the agent, including their timing and costs, and replace only the final decision with a fixed procedure. This procedure considers all treatments with confirmation results that pass approval review when replayed. It values the estimated benefit and harm of each treatment based on the prices specified in the brief to determine the estimated annual profit. The procedure then advances the treatment with the highest estimated annual profit if it is positive; otherwise, it stops the programme.

We run this procedure using three different sets of inputs: the agent's own final estimates, a standard analysis of the same participants, and the true effects. Compared to the agent's own decisions, which achieve 16.7\% of the oracle value, these achieve 17.2\%, 18.6\%, and 23.2\% of the oracle value, respectively. We can further modify the evidence considered. If we keep the agent's first single active arm confirmation study with the same participants, timing, and costs, but test the best treatment according to true effects, we achieve 31.1\% of the oracle value, compared to the RBD's 57.5\% on the same programmes. Even with perfect information at the final step and the optimal treatment in the same study, most of the value gap remains: the value is lost in which candidates agents study and how they size and time those studies, not in the final choice.

The only exception is Opus 5.5. Using its own estimates, the fixed procedure achieves 45.5\%, less than its actual value of 48.0\%. However, when using the true effects, the value increases to 60.8\%, against the RBD's 57.7\% on the same programmes. This indicates that the studies conducted by Opus 5.5 are comparable to the RBD's, but the agent's ability to value treatments based on these studies is lacking.

\begin{figure}[!htbp]\centering
\includegraphics[width=1.0\linewidth,height=0.9\textheight,keepaspectratio]{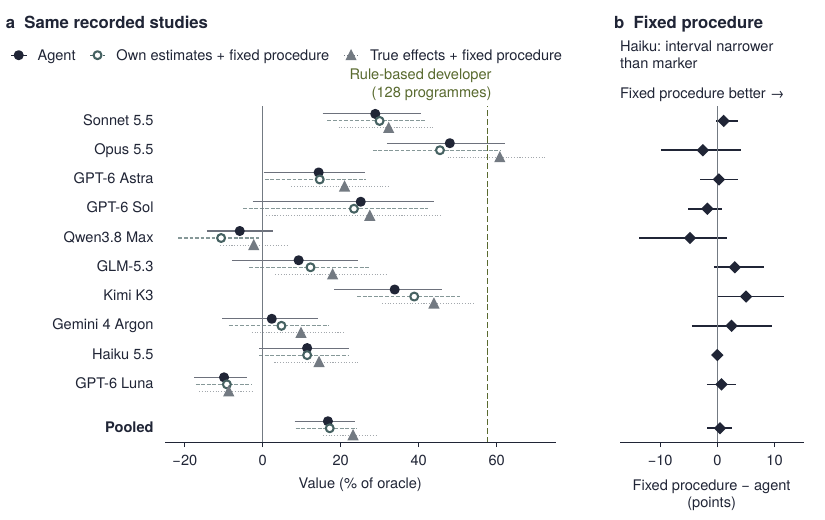}
\caption{Fixing the final decision based on an agent's own estimates provides minimal average improvement, while using true effects provides more and brings Opus 5.5 to parity with the rule-based developer. (a) Each row shows one agent's recorded studies valued as a percentage of oracle value. Black circle: agent's decisions. Open circle: fixed procedure using agent's own last reported estimates. Triangle: fixed procedure using true effects. Dashed line: rule-based developer using its own studies. Further right is better. (b) Difference between open circle and black circle, in points, with 95\% confidence intervals.}
\end{figure}

\subsection{Follow-up and horizon}

The value of treatments tested in short studies is overestimated, as the benefit may diminish and harm may accumulate after the study ends, particularly for intensified regimens. The programme brief specifies a time horizon for value assessment (224, 336, or 728 days, depending on the programme). At the median, the study duration leading to an advanced treatment is half the horizon. We value each advanced treatment in three different ways, using the prices from the brief: based on the agent's own estimates, based on the true effect at study follow-up, and based on the true effect at the horizon. The agent's own estimate is the effect estimate, with its interval, that the agent's saved analysis code reports at each results release; we use its last reported estimates of benefit and harm for the advanced treatment. The gap between the first two is due to errors in reading the study, while the gap between the last two is due to information that the study cannot capture.

Across 584 advanced treatments with usable estimates, the three valuations are US\$160 million, US\$141 million, and US\$98 million in annual profit. This indicates that US\$19 million of the discrepancy is due to study reading errors, while US\$43 million is due to the time horizon, which is more than double the former and is roughly equally split between diminishing benefit and accumulating harm. The discrepancy due to the time horizon is larger for intensified regimens (US\$59 million) compared to standard regimens (US\$31 million), so agents lose more by extrapolating beyond short studies than by misreading them, most of all for intensified regimens.

If we subtract the true discrepancy between the study follow-up and the horizon from the estimates of Opus 5.5 and rerun the fixed procedure, the value increases from 45.5\% to 53.9\%, a gain of 8.5 points (0.2 to 17.7). This is an upper bound, as in practice, an agent would need to estimate this discrepancy rather than having it provided.

\begin{figure}[!htbp]\centering
\includegraphics[width=1.0\linewidth,height=0.9\textheight,keepaspectratio]{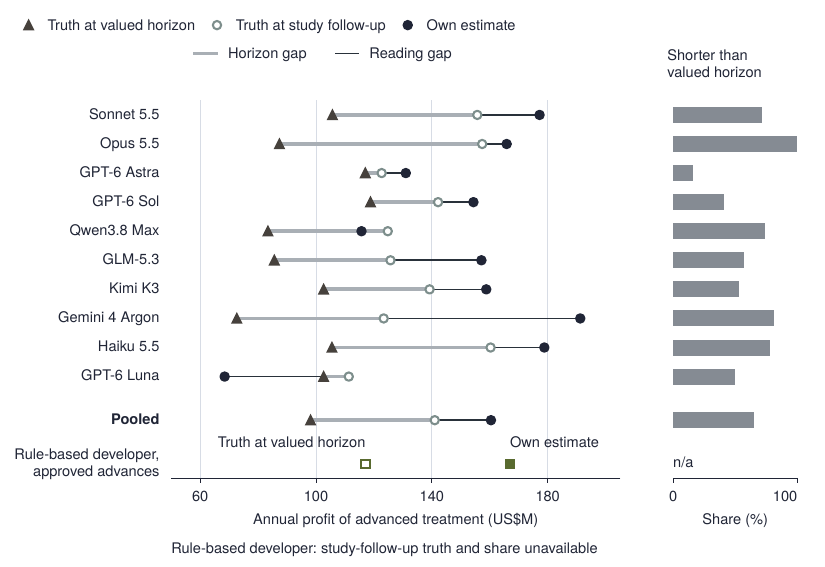}
\caption{Treatments have lower value at the valued horizon compared to the study end, and for most agents this gap is larger than their study reading error. Each row values the same advanced treatments three ways, as average yearly profit: using the agent's own estimates (black circle), true effect at the study's follow-up (open circle), and at the brief's horizon (triangle). Bars show the fraction of advances with follow-up shorter than the horizon. Line segments distinguish the reading gap (own estimate to truth at study follow-up) and horizon gap (study-follow-up truth to valued-horizon truth). Green squares identify the rule-based developer, which has no comparable study-follow-up value; its two marks show own estimate and valued-horizon truth.}
\end{figure}

\subsection{Price of harm}

Whether Opus 5.5 outperforms the RBD on fixed decisions depends on the price of harm. If the price of harm is scaled by a multiplier \(p\), a programme's value can be expressed as \(V(p) = B - pS - C\), where \(B\) is the discounted clinical value of the approved treatment, \(S\) is its cost of harm at the brief's price (both 0 if the treatment is not approved), \(C\) is the cost of studies and time, and \(p = 1\) is the price of harm in the brief. We fix all actions and approval outcomes and only vary the price of harm, so the value decreases linearly with the price of harm, by a tenth of \(S\) for every 10\% increase. The difference in value between two developers is also linear, \(\Delta V(p) = \Delta B - p\Delta S - \Delta C\), and the point at which this difference is 0 is the break-even price of harm, \((\Delta B - \Delta C)/\Delta S\).

Opus 5.5 generates US\$210.6 million more in clinical value per programme than the RBD, but incurs US\$213.5 million more in cost of harm and US\$30.4 million more in study and time costs. At the price of harm in the brief, it is outperformed by US\$33.2 million. If the price of harm were zero, it would be ahead by US\$180.2 million. The break-even is \((210.6 - 30.4)/213.5 = 0.84\), meaning that if the price of harm is 84\% (60 to 111) of the brief's price, the two would have the same value. This interval includes the brief's price. Opus 5.5 therefore gains more benefit than the RBD, but at the brief's price its extra cost of harm cancels that gain, because its estimates overstate the net value of intensified regimens at the valued horizon (Section 5.3). Opus 5.5 is the only agent that would outperform the RBD if the price of harm were zero, so it is not possible to close the gap for other agents by reducing the price of harm.

\begin{figure}[!htbp]\centering
\includegraphics[width=1.0\linewidth,height=0.9\textheight,keepaspectratio]{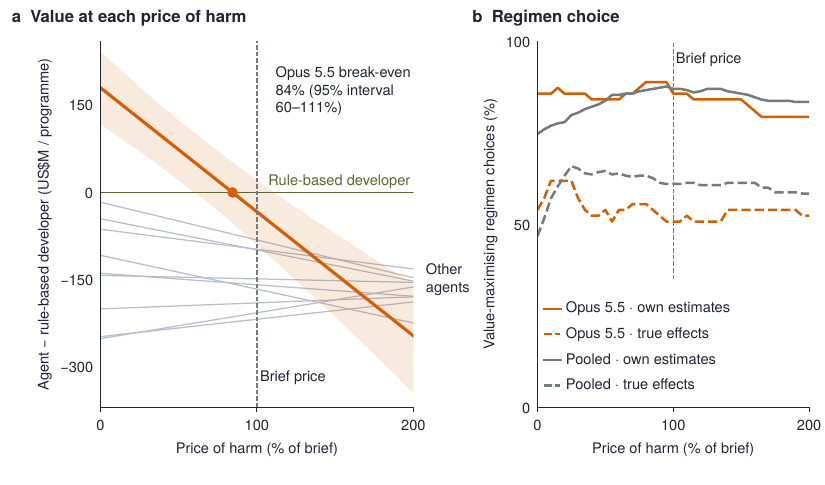}
\caption{With fixed actions, Opus 5.5 matches the rule-based developer at 84\% (60 to 111) of the brief's price of harm, and agents' regimen choices are more often correct based on their own numbers than on the truth. (a) Each line shows one agent's value minus the rule-based developer's value, in US\$ million per programme, at different multiples of the brief's price of harm; above zero means the agent is better. The highlighted line is Opus 5.5 with 95\% confidence band. (b) Among first confirmation studies with one active treatment and prior estimates for both regimens, the fraction of chosen regimens that maximise annual net value, based on own estimates (solid) or true effects at the valued horizon (dashed).}
\end{figure}

\begin{table}[!htbp]
\caption{Opus 5.5’s relative performance against the RBD is most sensitive to changes in the price of harm. Each row re-scores recorded actions under the stated prices; cells show values as percentages of oracle value. Gap is Opus 5.5 minus the RBD in points, with 95\% confidence intervals. The last column shows the rank correlation with values at the brief’s prices.}
\centering\scriptsize\setlength{\tabcolsep}{2pt}
\begin{tabular}{>{\raggedright\arraybackslash}p{0.207\linewidth}>{\raggedright\arraybackslash}p{0.165\linewidth}>{\raggedright\arraybackslash}p{0.087\linewidth}>{\raggedright\arraybackslash}p{0.228\linewidth}>{\raggedright\arraybackslash}p{0.233\linewidth}}
\toprule
\textbf{Prices} & \textbf{Rule-based developer} & \textbf{Opus 5.5} & \textbf{Opus 5.5 gap, points (95\% interval)} & \textbf{Rank correlation with brief's prices} \\
\midrule
Brief's prices & 57.7 & 48.0 & −9.7 (−26.9 to 6.3) & 1.00 \\
Profit halved, or cost doubled & 46.5 & 23.1 & −23.4 (−47.5 to −2.9) & 0.60 \\
Profit doubled, or cost halved & 61.7 & 56.9 & −4.8 (−20.1 to 9.1) & 0.91 \\
Price of harm halved & 57.1 & 75.4 & 18.3 (5.5 to 30.7) & 0.91 \\
Price of harm doubled & 59.7 & −49.0 & −108.7 (−176.1 to −63.9) & 0.38 \\
\bottomrule
\end{tabular}
\end{table}

The price of harm is important because the intensified regimen is sometimes the optimal choice. The oracle selects the intensified regimen for 39 out of 104 advanced treatments, accounting for 51\% of its value. The RBD never advances an intensified regimen and captures 42\% of the value on these programmes compared to 77\% on programmes where the oracle advances the standard regimen. Agents do consider harm when making decisions: based on their own estimates, the regimen they choose maximises value in 87\% of first confirmation studies. However, when considering true effects at the horizon, this drops to 61\%, so agents apply the brief's prices but misjudge how benefit and harm evolve by the horizon.

\subsection{Guardrails replayed on recorded programmes}

We apply each guardrail independently and re-score the recorded programmes. The guardrail that provides the most value checks public rules before submission, whereas guardrails based on the agents' own numbers provide little value or lose it.

The first guardrail checks if the agent submits a treatment for review with fewer than 300 completed treated participants. If the agent has enrolled at least 300 treated participants, we wait until they are completed and then replay the approval review on that day. This is exact and only costs additional time. If the agent has not enrolled 300 treated participants, we expand the last confirmation study in the agent's programme to include 300 treated participants from the unused pool. This is approximate because the exact site and screening dynamics are not modelled. We observe that 93 agent submissions violate this guardrail: 41 of them allow waiting, 15 of them allow expansion, and 37 of them exceed the budget or time limit. Overall, this guardrail improves value by 4.3 points (2.5 to 6.3), where 3.3 points are exact and 1.0 point is approximate.

The second guardrail triggers when the agent's latest 95\% confidence interval on the clinical benefit contains 0 and the agent still advances the treatment, which happens in 89 out of 584 advances (15\%). Enforcing this guardrail changes the value by −2.5 points (−3.9 to −1.3). Of the 89 programmes that trigger the guardrail, 27 pass approval review, and the value of these programmes is lost. For the other programmes that fail approval review, there is not much downside to submitting the programme for review as the agent has already paid for the studies.

Another guardrail replaces an intensified advance with the standard regimen if the agent's own estimates at the start of the confirmation phase favoured the standard regimen. This guardrail adds 0.9 points (0.2 to 1.8) of value. Out of 271 confirmation phase starts on intensified regimens where the agent has estimates on both regimens, the agent's own estimates favour the standard regimen in 32 cases and the true effects favour it in 180 cases. A guardrail based on the agent's own estimates cannot correct errors in those estimates.

\begin{figure}[!htbp]\centering
\includegraphics[width=1.0\linewidth,height=0.9\textheight,keepaspectratio]{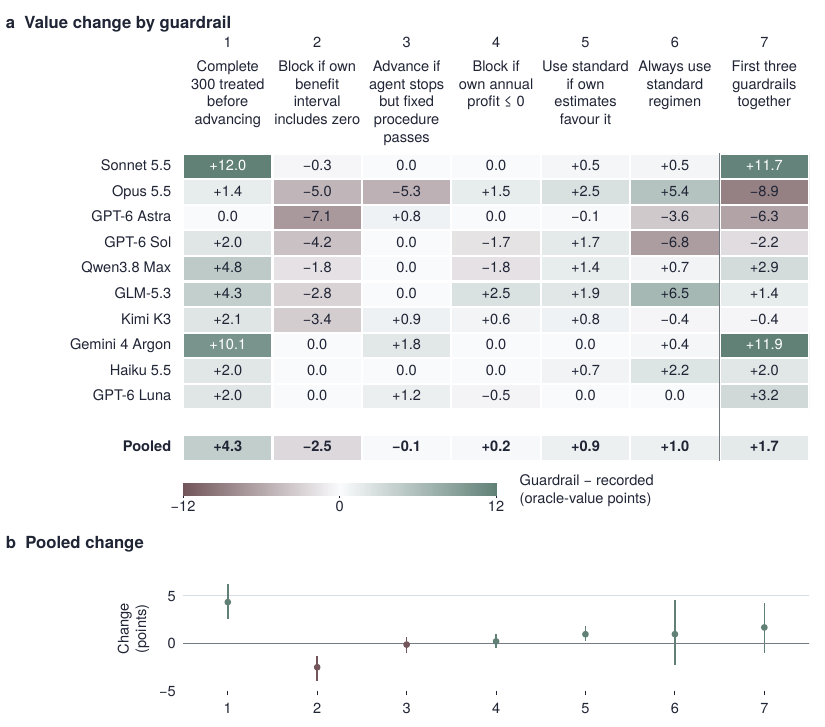}
\caption{Enforcing the 300-participant minimum adds the most value; blocking advances based on an agent's own confidence interval loses value. (a) Each cell shows the change in points of oracle value when replaying one guardrail alone on one agent's recorded programmes; positive values indicate gain and negative values indicate loss. Columns: complete 300 treated participants; block based on agent's own confidence interval; reverse a stop decision under the fixed procedure; block based on non-positive estimated annual profit; swap intensified to standard if agent's numbers favoured standard; swap all intensified advances to standard; the first three combined. (b) Pooled change, with 95\% confidence intervals; above zero is better.}
\end{figure}

\subsection{Advance decisions}

Agents are worse than the RBD at distinguishing programmes worth developing from those that are not. A programme is considered worth developing if the oracle achieves positive value from it. If a programme is not completed, it is considered not advanced. Agents advance 59\% of programmes worth developing and 32\% of programmes that are not. In comparison, the RBD advances 76\% of programmes worth developing and only 8\% of those that are not. Since the rate of advancing differs between developers, we compare them using d′ (Appendix D), which is calculated based on the hit rate of advancing programmes worth developing and the false positive rate of advancing programmes that are not. For agents, \(\Phi^{-1}(0.594) - \Phi^{-1}(0.315) \approx 0.24 - (-0.48) = 0.72\) (0.51 to 0.94). The RBD has a d′ of 2.2, which is 1.4 (1.0 to 2.3) better than the agents on the same set of programmes. This is partially due to the prior information the RBD has from its prespecified statistical model.

\begin{figure}[!htbp]\centering
\includegraphics[width=1.0\linewidth,height=0.9\textheight,keepaspectratio]{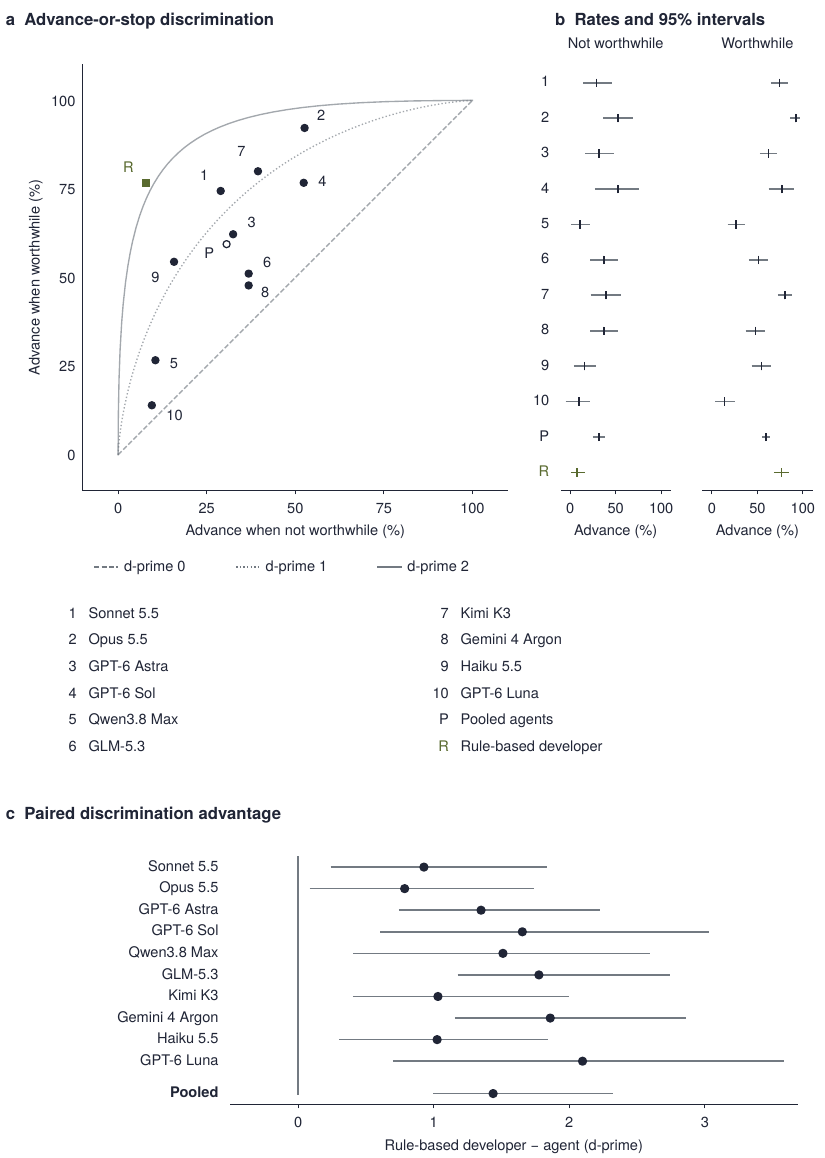}
\caption{Agents are worse than the rule-based developer at distinguishing programmes worth developing from those that are not. (a) Sensitivity against false-positive rate; the open point pools all agents and R is the rule-based developer. Curves show d′ of 0, 1 and 2. (b) The same rates and horizontal 95\% intervals, separated by model. (c) Paired rule-minus-agent d′ with 95\% intervals. Worthwhile means positive oracle value.}
\end{figure}

\FloatBarrier
\section{Analysis and development value}
\setcounter{table}{0}\setcounter{figure}{0}\renewcommand{\thetable}{\Alph{section}.\arabic{table}}\renewcommand{\thefigure}{\Alph{section}.\arabic{figure}}

More accurate analysis generally goes with higher value, but analysis accuracy is not the sole determinant of development value.

\subsection{Across models, lower error goes with higher value}

We compare the 10 models that have results on both the TrialEval and TrialDev suites. We rank the 10 models on the three TrialEval metrics and the five TrialDev metrics. We then use Spearman correlation to compare the ranks of the models on each TrialEval metric to their ranks on each TrialDev metric. Since both lower analysis error and higher development value are better, we expect to see a negative correlation between analysis error and development value. We find one (ρ = −0.64), and also between serious misses and development value (ρ = −0.72), between analysis error and clinical value (ρ = −0.64), and between analysis error and approvals (ρ = −0.61). This indicates that the models that avoid serious misses tend to be the most valuable.

For cost of harm and study cost, for which higher values are worse, we see negative correlations of ρ = −0.48 and ρ = −0.55, respectively, indicating that more accurate models tend to have higher costs of harm and study costs. The coverage metric does not correlate with development value (ρ = −0.03) because all 10 models have similar coverage ranging from 92\% to 95\%.

\begin{table}[!htbp]
\caption{Models with lower TrialEval error achieve higher TrialDev value, with a correlation of −0.64. Each cell shows the Spearman correlation between one TrialEval metric (row) and one TrialDev outcome (column) across the 10 models that ran both evaluation suites, with a 95\% confidence interval.}
\centering\scriptsize\setlength{\tabcolsep}{2pt}
\begin{tabular}{>{\raggedright\arraybackslash}p{0.134\linewidth}>{\raggedright\arraybackslash}p{0.155\linewidth}>{\raggedright\arraybackslash}p{0.155\linewidth}>{\raggedright\arraybackslash}p{0.155\linewidth}>{\raggedright\arraybackslash}p{0.155\linewidth}>{\raggedright\arraybackslash}p{0.155\linewidth}}
\toprule
\textbf{TrialEval measure} & \textbf{Value} & \textbf{Clinical value} & \textbf{Cost of harm} & \textbf{Study cost} & \textbf{Approvals} \\
\midrule
Average error & −0.64 (−0.82 to −0.33) & −0.64 (−0.83 to −0.43) & −0.48 (−0.77 to −0.25) & −0.55 (−0.78 to −0.33) & −0.61 (−0.84 to −0.43) \\
Serious misses & −0.72 (−0.85 to −0.33) & −0.72 (−0.87 to −0.39) & −0.61 (−0.81 to −0.26) & −0.57 (−0.84 to −0.27) & −0.70 (−0.88 to −0.41) \\
Coverage & −0.03 (−0.39 to 0.44) & 0.01 (−0.28 to 0.48) & −0.32 (−0.53 to 0.26) & 0.15 (−0.05 to 0.59) & 0.04 (−0.27 to 0.50) \\
\bottomrule
\end{tabular}
\end{table}

\subsection{Within a programme, an agent's own accuracy does not predict value}

Within a programme, the accuracy of an agent's estimates of the clinical outcomes has low correlation with the net development value it achieves (ρ = +0.03 on aggregate, ρ = −0.05 after controlling for the programme and model). On the set of programmes where the agents have the most accurate analysis, they only capture 23\% of the oracle's value. In contrast, the RBD captures 82\% of the oracle's value on the same set of programmes. How accurately an agent reads its own studies says little about how well it decides.

\begin{table}[!htbp]
\caption{An agent’s value does not increase as its own estimates become more accurate. Each row represents one quartile of programmes sorted by the agent’s error on clinical outcomes. Cells show agent and RBD values as percentages of oracle value, with 95\% confidence intervals, and the mean difference in US\$M.}
\centering\scriptsize\setlength{\tabcolsep}{2pt}
\begin{tabular}{>{\raggedright\arraybackslash}p{0.141\linewidth}>{\raggedright\arraybackslash}p{0.182\linewidth}>{\raggedright\arraybackslash}p{0.171\linewidth}>{\raggedright\arraybackslash}p{0.243\linewidth}>{\raggedright\arraybackslash}p{0.183\linewidth}}
\toprule
\textbf{Mean own error} & \textbf{Programmes} & \textbf{Agents, \% of oracle} & \textbf{Rule-based developer, \% of oracle} & \textbf{Agent minus RBD, US\$M} \\
\midrule
0.11 & 232 & 23 (6 to 35) & 82 (73 to 90) & −160 \\
0.24 & 232 & 21 (5 to 32) & 71 (61 to 81) & −169 \\
0.49 & 231 & 12 (−4 to 25) & 63 (52 to 73) & −165 \\
2.84 & 232 & 32 (14 to 45) & 33 (7 to 57) & −7 \\
\bottomrule
\end{tabular}
\end{table}

\FloatBarrier
\section{The initial submission and how the updated release differs}
\setcounter{table}{0}\setcounter{figure}{0}\renewcommand{\thetable}{\Alph{section}.\arabic{table}}\renewcommand{\thefigure}{\Alph{section}.\arabic{figure}}

\subsection{The initial submission graded against accepted methods}

In the initial submission, we benchmarked 6 models on 400 TrialEval items and 60 TrialDev programme runs for each model. We graded outputs against a predefined target, which partially depended on the model using an accepted method from a list of our own, which we committed to having domain experts adjudicate. Under strict grading (the model uses an accepted method and stays within tolerance) no model achieved greater than 24.5\% and performance was sensitive to which methods counted as accepted. On TrialDev, the two models that completed all programmes made the correct decision on whether to stop or proceed less than a third of the time. Applied to the updated release, strict grading marked as incorrect 28\% of answers that were as accurate as the process-aware analysis (Appendix G.3), and domain experts advised us to fix the estimand for each task and score against a simulated truth, as in simulation studies. The updated release therefore scores every method against the true effect.

\subsection{The updated release scores against the truth}

The updated release fixes the clinical question for each task, allows any method, and scores the saved analysis based on its error with respect to the true effect on replication trials. On TrialDev, the updated release scores the value (benefit, harm, and cost) of each decision as a fraction of oracle value and against the RBD.

The scores from the initial submission are not on the same scale. They grade outputs against a target where higher is better and missing outputs or incomplete programmes are counted as failures. On the updated release, error is measured in standard errors where lower is better, and value ranges from 0\% value for stopping immediately to 100\% for the oracle, but can fall outside this range.

\begin{table}[!htbp]
\caption{The initial submission used target-based grading: no model achieved more than 24.5\% strict grading on items. Each row shows one model, with 95\% confidence intervals; higher is better. Family counts items completed with the accepted analysis family.}
\centering\scriptsize\setlength{\tabcolsep}{2pt}
\begin{tabular}{>{\raggedright\arraybackslash}p{0.104\linewidth}>{\raggedright\arraybackslash}p{0.125\linewidth}>{\raggedright\arraybackslash}p{0.089\linewidth}>{\raggedright\arraybackslash}p{0.067\linewidth}>{\raggedright\arraybackslash}p{0.055\linewidth}>{\raggedright\arraybackslash}p{0.134\linewidth}>{\raggedright\arraybackslash}p{0.134\linewidth}>{\raggedright\arraybackslash}p{0.078\linewidth}>{\raggedright\arraybackslash}p{0.087\linewidth}}
\toprule
\textbf{Model} & \textbf{TrialEval all-item \% (95\% CI)} & \textbf{Completed} & \textbf{Family} & \textbf{Strict} & \textbf{Same-estimand} & \textbf{TrialDev score (95\% CI)} & \textbf{Complete} & \textbf{Stop timing right} \\
\midrule
GPT-5.4 & 38.2 (33.7--42.7) & 400 & 215/400 & 24.50 & 26.50 & 0.429 (0.389--0.471) & 60 & 23\% \\
Gemini 3.1 Pro Preview & 30.5 (26.3--34.6) & 374 & 155/374 & 21.50 & 23.75 & 0.348 (0.297--0.398) & 51 & 25\% \\
GLM-5.1 & 25.4 (21.5--29.7) & 263 & 132/263 & 19.75 & 20.75 & 0.138 (0.082--0.197) & 17 & 32\% \\
Opus 4.7 & 25.2 (21.2--29.3) & 392 & 201/392 & 14.75 & 16.50 & 0.417 (0.381--0.454) & 60 & 27\% \\
Qwen3.6 Plus & 15.6 (12.3--19.0) & 116 & 76/116 & 13.75 & 13.75 & 0.081 (0.033--0.135) & 9 & 45\% \\
Kimi K2.6 & 13.1 (9.9--16.4) & 106 & 56/106 & 10.00 & 10.25 & 0.252 (0.190--0.316) & 31 & 47\% \\
\bottomrule
\end{tabular}
\end{table}

\subsection{Accepted-method grading misses accurate analyses}

We applied strict grading on the updated release for the 22 models that we could re-grade and compared the grades to the error on replication trials. Overall, 28\% of the outputs graded as incorrect were as accurate as the process-aware analysis.

\end{document}